\documentclass[10pt,twocolumn,letterpaper]{article}

\usepackage[pagenumbers]{cvpr} 

\usepackage{amsmath}
\usepackage{colortbl}
\usepackage{capt-of}
\usepackage[dvipsnames,table,xcdraw]{xcolor}
\usepackage{xspace}
\usepackage{enumitem}
\usepackage{amsmath,amssymb,amsbsy,amsfonts,dsfont,pifont,bm,bbm,mathrsfs,mathtools,nicefrac}
\usepackage{algorithm,algpseudocode,listings}
\usepackage{booktabs,multirow,adjustbox,diagbox,threeparttable,tabularray}
\usepackage{multirow}
\usepackage{overpic}
\usepackage{indentfirst}

\newcommand{\tocite}[1]{{\color{red} [TO CITE]}}

\definecolor{NearOOD}{HTML}{E6ECE3}
\definecolor{FarOOD}{HTML}{ffefe0}

\definecolor{cvprblue}{rgb}{0.21,0.49,0.74}
\usepackage[pagebackref,breaklinks,colorlinks,citecolor=cvprblue]{hyperref}
\usepackage{wrapfig}
\usepackage[capitalize]{cleveref}  
\crefname{section}{Sec.}{Secs.}
\Crefname{section}{Section}{Sections}
\crefname{table}{Tab.}{Tabs.}
\Crefname{table}{Table}{Tables}
\crefname{figure}{Fig.}{Figs.}
\Crefname{figure}{Figure}{Figures}
\crefname{equation}{Eq.}{Eqs.}
\Crefname{equation}{Equation}{Equations}
\newcommand\blfootnote[1]{%
  \begingroup
  \renewcommand\thefootnote{}\footnote{#1}%
  \addtocounter{footnote}{-1}%
  \endgroup
}

\def\id{\mathcal{D}_{\text{ID}}}
\def\idlabel{\mathcal{Y}_{\text{ID}}}
\def\idtest{\mathcal{T}_{\text{ID}}}

\def\oodtest{\mathcal{T}_{\text{OOD}}}
\def\nearoodtest{\mathcal{T}_{\text{OOD}}^\texttt{near}}
\def\faroodtest{\mathcal{T}_{\text{OOD}}^\texttt{far}}

\def\csIDtest{{\mathcal{T}_{\text{csID}}}}

\def\paperID{2182}
\def\confName{CVPR}
\def\confYear{2024}

\usepackage[marginal]{footmisc}
\renewcommand{\thefootnote}{}

\begin{document}

\title{Likelihood-Aware Semantic Alignment for Full-Spectrum \\ Out-of-Distribution Detection}

\author{Fan Lu$^{1,*,\ddagger}$\qquad~Kai Zhu$^{1,2,*}$\qquad~Kecheng Zheng$^{3}$\qquad~Wei Zhai$^{1}$\qquad~Yang Cao$^{1,4,\dagger}$\\
{$^{1}$~University of Science and Technology of China} \quad
{$^{2}$~Alibaba Group} \quad
{$^{3}$~Ant Group} \\
{$^{4}$~Institute of Artificial Intelligence, Hefei Comprehensive National Science Center}\\
\small{\texttt{\{lufan@, zkzy@\}mail.ustc.edu.cn}} \qquad
\small{\texttt{zkccloud@gmail.com}} \qquad
\small{\texttt{\{wzhai056@, forrest@\}ustc.edu.cn}}
  }

\maketitle

\blfootnote{\noindent $*$Co-first Author. $\dagger$Corresponding Author. \\ $\ddagger$Work done during an internship at Ant Group.}

\begin{abstract}

%
Full-spectrum out-of-distribution (F-OOD) detection aims to accurately recognize in-distribution (ID) samples while encountering semantic and covariate shifts simultaneously.
However, existing out-of-distribution (OOD) detectors tend to overfit the covariance information and ignore intrinsic semantic correlation, inadequate for adapting to complex domain transformations.
To address this issue, we propose a Likelihood-Aware Semantic Alignment (LSA) framework to promote the image-text correspondence into semantically high-likelihood regions. LSA consists of an offline Gaussian sampling strategy which efficiently samples semantic-relevant visual embeddings from the class-conditional Gaussian distribution, and a bidirectional prompt customization mechanism that adjusts both ID-related and negative context for discriminative ID/OOD boundary.
%
%
%
Extensive experiments demonstrate the remarkable OOD detection performance of our proposed LSA especially on the intractable Near-OOD setting, surpassing existing methods by a margin of $15.26\%$ and $18.88\%$ on two F-OOD benchmarks, respectively. Code will be available at \href{https://github.com/LuFan31/LSA}{https://github.com/LuFan31/LSA}.

\end{abstract}

\section{Introduction}\label{sec:intro}

\begin{figure}[t]
\small
\centering
\begin{overpic}[width=1\linewidth]{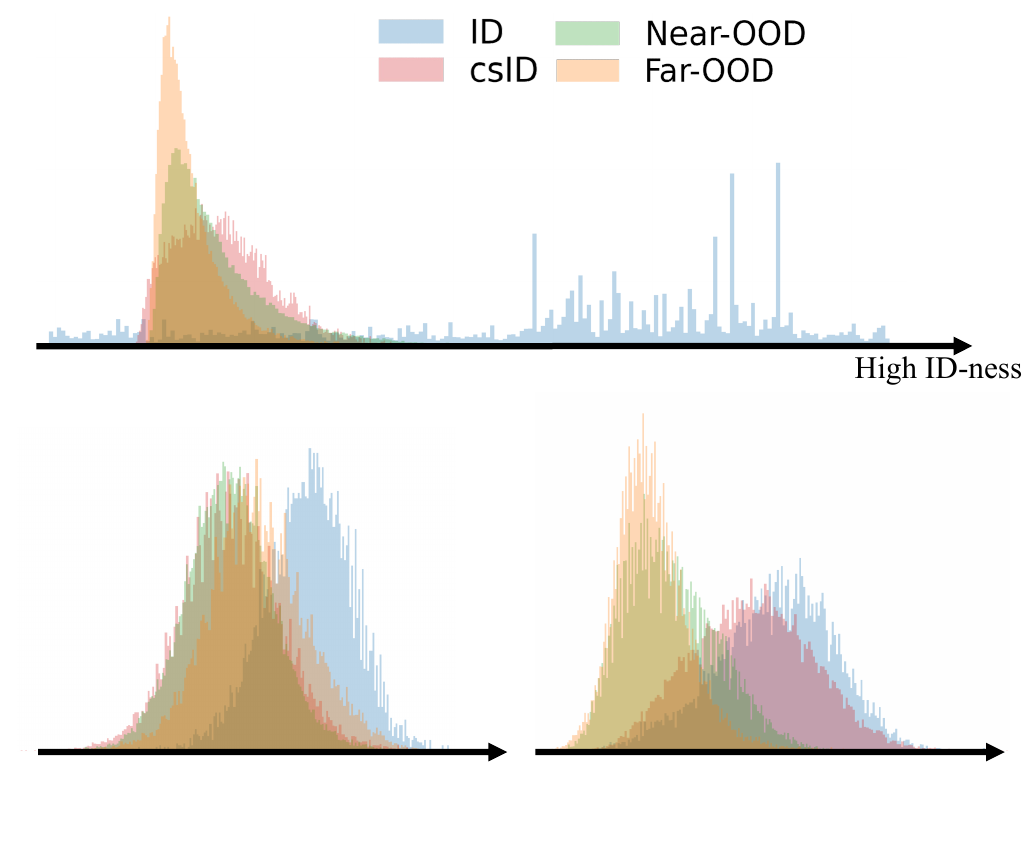}
        \put(18, 45){(a) Energy score of EBO on F-OOD}
        \put(8, 5.5){(b) Energy score of}
        \put(12, 0.5){CLIP on F-OOD}
        \put(57, 5.5){(c) Energy core of}
        \put(54, 0.5){\textbf{LSA (ours)} on F-OOD}
\end{overpic}
\vspace{-1.3em}
\caption{\textbf{Different energy distributions on F-OOD.}  \textbf{(a)} Existing OOD methods such as EBO \cite{energyood} confuse the semantic consistency of F-OOD on energy distribution, where a part of ID data obtains lower ID-ness than OOD and csID overlaps with OOD significantly. \textbf{(b)} Directly incorporating CLIP into F-OOD task does not eliminate the severe overlap between csID and OOD samples despite the more concentrated values of the ID cluster. \textbf{(c)} Our proposed LSA maintains consistent distribution between csID and ID data on energy score, which benefits from enhancing semantic alignment during context optimization. The energy scores in (b) and (c) are computed using image-text cosine similarity.}
\vspace{-8pt}
\label{fig:Fig1}
\end{figure}

\begin{figure}[t]
\small
\centering
\begin{overpic}[width=1\linewidth]{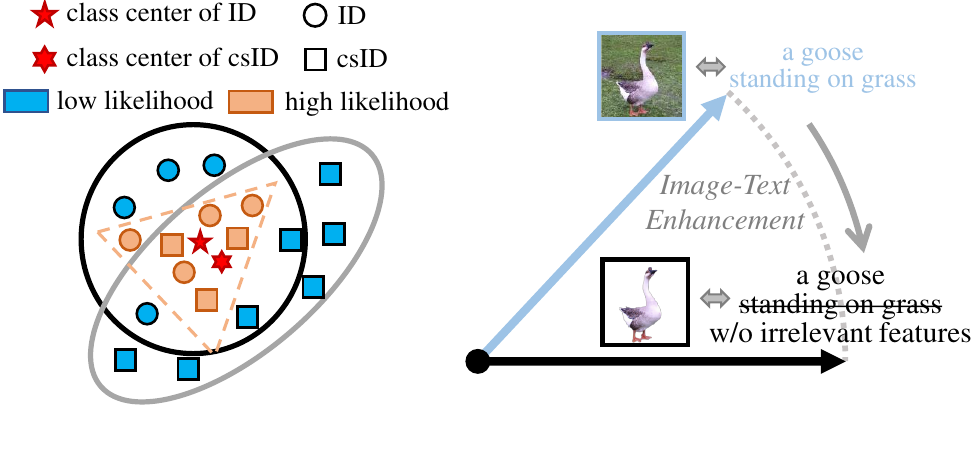}
        \put(2, 4){(a) Feature distribution}
        \put(50, 4){(b) Optimization direction}
\end{overpic}
\vspace{-2.1em}
\caption{\textbf{Motivation of the proposed method.} \textbf{(a)} Compared to the discrepancy within the global distribution, high-likelihood features (\textit{i.e.}, ones closer to the class center) from ID and csID clusters are more compact and consistent, which helps resist the interference of covariate shift. \textbf{(b)} Considering that the feature space of the CLIP-based model is jointly determined by the image-text pairs, our method simultaneously highlights the semantically high-likelihood visual regions and reduce the fitting to ID-irrelevant contexts.}
\vspace{-8pt}
\label{fig:Fig2}
\end{figure}

Deep visual models \cite{he2016deep, dosovitskiy2020image} have demonstrated remarkable performance in closed-set environments. However, their performance significantly deteriorates when faced with out-of-distribution (OOD) samples in real-world scenarios, such as input from unknown classes \cite{deepfool}. 
To enhance the deployment security, OOD detection has received increasing research interest recently  \cite{medical,geiger2012arewereadyfor,hendrycks2016baseline,du2022vos}.

A rich line of OOD detection methods \cite{hendrycks2016baseline,hendrycks18oe,odin,energyood,mcd} involves one in-distribution (ID) dataset in training while regarding all other datasets as OOD, where models tend to overfit the low-level covariate shift while ignoring the inherent semantic correlation across different datasets \cite{yang2021semantically,yang2023full,zhang2023openood}. For instance, arbitrarily considering pet dogs and cartoon dogs as separate categories is unjust, as both share the semantic concept of `dog', despite displaying distinct covariance information. Disregarding this fact in realistic scenarios can give rise to irreparable interference.

To comprehensively evaluate covariate shift and semantic shift simultaneously, the realistic F-OOD detection benchmarks are proposed in OpenOODv1.5 \cite{zhang2023openood}. F-OOD benchmarks split OOD data to Near-OOD and Far-OOD based on their degrees of covariate shift, and introduce covariate-shifted ID (csID) data from training-agnostic domains that retain ID concepts. Regrettably, the experimental evidence \cite{zhang2023openood} proves that numerous existing OOD detectors have experienced notable performance decline on the F-OOD benchmarks. We analyze this phenomenon in \cref{fig:Fig1}\textcolor{red}{(a)} and reveal that classical OOD detectors like EBO \cite{energyood} fail to establish effective discrimination within a broad semantic scope, particularly when it comes to differentiating csID from OOD samples, let alone aggregating ID-related (\textit{i.e.}, csID and ID) features uniformly.

Although several methods \cite{zoc,mcm,zeroMCMlocal,miyai2023locoop} leverage the perceptual profundity of CLIP model \cite{maskclip,masksegmentation,OVSeg} to enhance the aggregation of fine-grained semantics, we observe that naively incorporating it into F-OOD still fails to distinguish between OOD and csID clusters as shown in \cref{fig:Fig1}\textcolor{red}{(b)}, which is mainly due to the significant overlap of low-confidence regions.
Delving into the likelihood characteristics of CLIP-based image space, we discovered in \cref{fig:Fig2}\textcolor{red}{(a)} that high-likelihood ID or csID samples close to respective class centers share a more compact and consistent distribution than the global semantic data, which contributes to resilience against interference from covariate shift.
Motivated by this, we consider enhancing the alignment of high-likelihood regions based on the above statistical characteristics. Moreover, as shown in \cref{fig:Fig2}\textcolor{red}{(b)}, while irrelevant attributes or relationships are filtered out for heightened image features, the matching textual context inevitably experiences drift in both the ID and OOD regions, consequently affecting the semantic-specific discriminability.

To synergistically advance the aggregation of high-likelihood features, we propose a likelihood-aware semantic alignment (LSA) framework that can be divided into visual and textual perspectives. For the design of the image branch, we construct an offline Gaussians sampling strategy, effectively modeling the whole probability density. Specifically, we first represent the ID images embeddings as a class-conditional Gaussian distribution utilizing the feature statistics of each class. Afterwards, we capture samples with the highest and lowest probability density from the feature Gaussians as ID semantic-relevant visual regions and OOD regularization regions, respectively.
For the design of textual prompts, we present a bidirectional prompt customization mechanism, flexibly adjusting the ID/OOD boundary. Firstly, we enrich specific prompts for each ID class name, covering the diversity of ID foreground objects. Moreover, we portray unknown OOD prompts from scratch by aligning with the above OOD regularization regions, which serves as the `negative anchor' role to enlarge the discrepancy between ID and OOD semantics in text space.  \cref{fig:Fig1}\textcolor{red}{(c)} illustrates that our method maintains consistency for data with the same semantic, benefiting from our implementation strategies.



Our contributions are summarized as follows: \textbf{1)} We explore and unleash the potential of the CLIP model in terms of perceiving semantic and covariate shifts in OOD detection. \textbf{2)} We present a customized prompt tuning method for the F-OOD task named LSA, which enhances the image-text alignment with semantically high-likelihood regions. \textbf{3)} Extensive experiments demonstrate that our proposed method achieves state-of-the-art performance on the realistic F-OOD benchmarks.
\section{Related Work}\label{sec:related}

\definecolor{flamecolor}{RGB}{233,161,72}
\definecolor{snowflakecolor}{RGB}{91,157,219}
\begin{figure*}
\centering
\small
\begin{overpic}[width=1\linewidth]{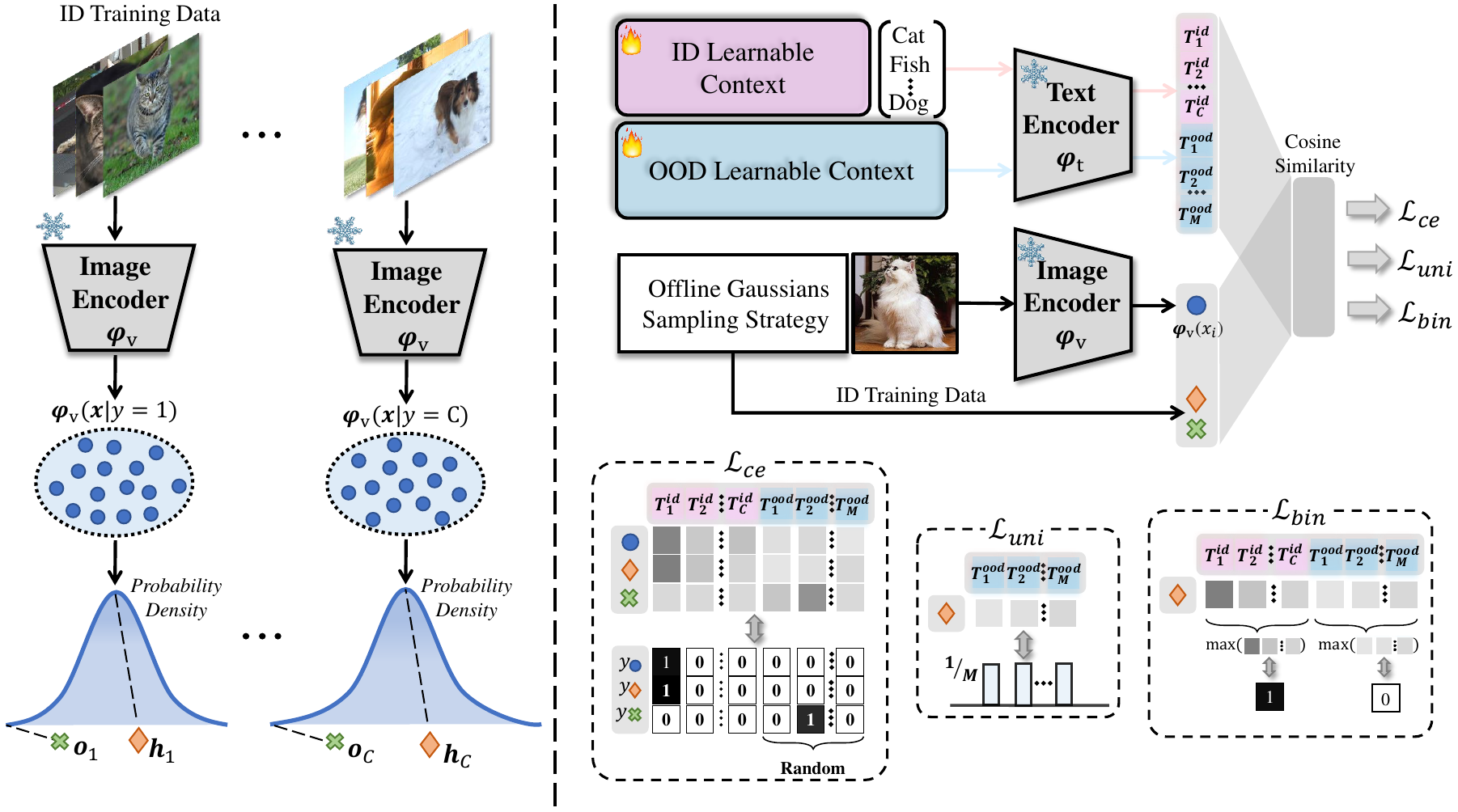}
        \put(4, 0.5){{(a) Offline Gaussians Sampling Strategy}}
        \put(63, 0.5){{(b) Training Pipeline}}
\end{overpic}
\vspace{-8pt}
  \caption{\textbf{Overall framework of our proposed LSA.} \textbf{(a)} The offline class-conditional Gaussian distribution is first modeled with the statistics of ID embeddings, then we select samples with the highest and lowest probability density in the Gaussians as ID semantic-relevant regions $\boldsymbol{h}$ and OOD regularization regions $\boldsymbol{o}$, respectively. \textbf{(b)} ID and OOD contexts are optimized through three losses. $\mathcal{L}_{ce}$ align ID context with ID image embedding and $\boldsymbol{h}$, and bring OOD context and $\boldsymbol{o}$ close, where every $\boldsymbol{o}$ obtains a random OOD label. $\mathcal{L}_{uni}$ unify the similarity of $\boldsymbol{h}$ with OOD text embedding. Meanwhile, a binary sigmoid loss $\mathcal{L}_{bin}$ is utilized to encourage ID and OOD textual embeddings to further separate and enlarge the ID/OOD semantic discrepancy. \textbf{\textcolor{flamecolor}{Flames}} and \textbf{\textcolor{snowflakecolor}{Snowflakes}} refer to learnable and frozen parameters, respectively.}
  \label{fig:method}
  \vspace{-5pt}
\end{figure*}

\subsection{Out-of-Distribution Detection}
OOD samples in OOD detection tasks belong to different categories from the known ones and OOD detectors should concentrate on identifying semantic shift. Several classic works consider one dataset as ID and define all other datasets as OOD \cite{hendrycks2016baseline,odin,energyood,hendrycks18oe,mcd,maxlogits}. However, OOD detection models under this benchmark tend to overfit the low-level covariate shift and disregard the high-level semantic discrepancy between samples, conflicting with the goal of OOD detection. 

To overcome this issue, more realistic settings including SCOOD benchmarks \cite{yang2021semantically} and F-OOD benchmarks \cite{yang2023full,zhang2023openood} are proposed. SCOOD benchmarks require models to distinguish ID and OOD samples mixed in an unlabeled dataset, UDG \cite{yang2021semantically} and ET-OOD \cite{lu2023uncertainty} respectively apply K-means clustering and energy-based optimal transport to make models understand the semantic knowledge hidden in the unlabeled dataset. The more challenging but practical F-OOD benchmarks \cite{yang2023full,zhang2023openood} introduce ID samples with changes in appearances like style, lighting or viewpoint, which can be easily misclassified as OOD just owing to the covariate shift. Unfortunately, OpenOODv1.5 \cite{zhang2023openood} has proved that dozens of existing OOD detectors are so sensitive to covariate shift that they encounter comprehensively significant performance degradation in the F-OOD detection problem.
All of these indicate that the fundamental challenge of OOD detection tasks, which is how to be robust to covariate shift and fully focus on the semantic shift between ID/OOD samples, necessitates increased research efforts. In this work, we explore how to highlight the semantic knowledge in the F-OOD detection task, and propose the LSA which enhances the image-text alignment with semantically high-likelihood regions. 

\subsection{Pre-trained Vision-Language Models}
Pre-trained vision-language models (\textit{e.g.}, CLIP \cite{radford2021learning}) align the rich multi-modal representations using large-scale image-text pairs during training instead of learning with only image supervision, which enables CLIP to perform open-world image recognition tasks such as open-vocabulary semantic segmentation and OOD detection. Several CLIP-based open-vocabulary semantic segmentation works \cite{maskclip,OVSeg,masksegmentation} have successfully enhanced the perception and localization of target semantic regions by aligning semantic regions in images and specified textual concepts. Previous studies on CLIP-based OOD detection primarily explore the performance of zero-shot OOD detection with CLIP \cite{zoc,mcm,zeroMCMlocal} and the impact of classic CLIP fine-tuning methods \cite{coop,cocoop,zhang2021tip} on OOD detection performance \cite{clipftood}. 
Recent works including CoOp \cite{coop}, CoCoOp \cite{cocoop} and MaPLe \cite{khattak2023maple}, extend prompt tuning originating from NLP \cite{jiang2020can,autoprompt} to computer vision and aim to fine-tune foundation models to adapt to new tasks in a parameter-efficient way.
Inspired by this, LoCoOp \cite{miyai2023locoop} applies prompt learning with CLIP in OOD detection by regarding ID-irrelevant regions as OOD regularization and pushing them away from the ID class text embedding. 
However, existing CLIP-based OOD detection methods ignore the semantic-relevant visual knowledge (i.e., foreground objects) and align them with the learnable ID class-specific context. In this paper, we propose LSA, which is a customized prompt tuning method for F-OOD, to unleash the ability of CLIP to focus on semantic shift.

\section{Method}\label{sec:method}

\subsection{Problem Statement}
During training in the F-OOD task, we can only access the in-distribution (ID) training set $\id$, and the corresponding label set is $\idlabel$. Additional to the ID testing set $\idtest$, F-OOD introduces the data from different domains compared to $\id$ while retaining ID semantics as covariate-shifted ID (csID) set formed as $\csIDtest$ during test time, where $\forall(x_{\text{csID}},y_{\text{csID}})\sim\csIDtest,y_{\text{csID}}\in\idlabel$. Moreover, the OOD testing set $\oodtest$ from the semantics not overlapping with $\idlabel$ is split into Near-OOD ($\nearoodtest$) and Far-OOD ($\faroodtest$) which represent two different levels of covariate shift. We desire to identify data from $\oodtest$ as OOD while classifying samples from $\idtest$ and $\csIDtest$ correctly.


\subsection{Overall Framework}
The framework of our LSA is presented in \cref{fig:method}. We first model the ID visual embeddings as an offline class-conditional Gaussian distribution in \cref{fig:method}\textcolor{red}{(a)}, and then select regions with the highest/lowest probability density from each ID class Gaussians, which will be involved in training. During training shown in \cref{fig:method}\textcolor{red}{(b)}, we apply $\mathcal{L}_{ce}$ and $\mathcal{L}_{uni}$ to learn ID and OOD contexts through aligning them with the corresponding visual regions sampled from Gaussians, and $\mathcal{L}_{bin}$ further widen the discrepancy between ID and OOD semantics in text space.

\subsection{Offline Gaussians Sampling Strategy}\label{subsec:off-line}
To obtain ID semantic-relevant regions and OOD regularization regions for alignment with class-specific contexts without extra segmentation models \cite{sam} or generative models \cite{gan}, we represent the ID visual embeddings as a class-conditional Gaussian distribution and sample distinguishable likelihood regions from the feature Gaussians. 

We assume the probability density of visual embeddings follow the class-conditional Gaussian distribution based on the hypothesis made in \cite{lee2018simple}: $p_\theta(\boldsymbol{\varphi}_v(\boldsymbol{x}) | y=c)=\mathcal{N}(\boldsymbol\mu_{c}, \boldsymbol\Sigma_{c})$, where $\boldsymbol{\varphi}_v$ is the visual encoder of CLIP, $\boldsymbol\mu_{c}$ and $ \boldsymbol\Sigma_{c}$ are the mean and covariance matrix of the $c$-$th$ ID class, respectively. To parameterize the ID class-conditional Gaussians with the feature statistics, we calculate the class mean $\widehat{\bm\mu}_{c}$ and covariance $\widehat{\boldsymbol{\Sigma}}_{c}$ as:
\vspace{-0.3em}
\begin{equation}
    \widehat{\bm\mu}_{c}=\frac{1}{N_{c}} \sum_{y_{i}=c} \boldsymbol{\varphi}_v(\*x_i),
    \label{eq:mean}
\vspace{-0.3em}
\end{equation}

\vspace{-1.45em}
\begin{equation}
    \widehat{\boldsymbol{\Sigma}}_c=\frac{1}{N_{c}} \sum_{y_{i}=c}\left(\boldsymbol{\varphi}_v(\*x_i)-\widehat{\bm\mu}_{c}\right)\left(\boldsymbol{\varphi}_v(\*x_i)-\widehat{\bm\mu}_{c}\right)^{\top},
    \label{eq:cov}
\vspace{-0.35em} 
\end{equation}
where $x_i \in \id$ and $N_c$ is the number of samples in ID class $c$. Then we can flexibly select regions from the estimated Gaussians based on the likelihood. Specifically, we sample $N$ data from the above Gaussians for each ID class and obtain a class-specific probability density set $\{f(\boldsymbol{a}_{c}^1),f(\boldsymbol{a}_{c}^2),...,f(\boldsymbol{a}_{c}^N)\}$, where $f(\cdot)$ is the function of Gaussian probability density, and the smaller probability density corresponds to the larger distance from the mean (\textit{i.e.}, class center). As a result, we respectively capture data with the highest and lowest probability density as ID semantic-relevant and OOD regularization regions:
\vspace{-0.3em}
\begin{equation}
    \boldsymbol{h}_{c}=\boldsymbol{a}_{c}^i, \ \text{where} \ \ i=\mathop{\arg\max}\limits_{n} \ f(\boldsymbol{a}_{c}^n),
    \label{eq:cen}
    \vspace{-0.3em}
\end{equation}

\vspace{-1.1em}
\begin{equation}
    \boldsymbol{o}_{c}=\boldsymbol{a}_{c}^j, \ \text{where} \ \ j=\mathop{\arg\min}\limits_{n} \ f(\boldsymbol{a}_{c}^n),
    \label{eq:ood}
\vspace{-0.35em}
\end{equation}
where $\boldsymbol{h}_{c}, \boldsymbol{o}_{c} \in \mathbb{R}^{D}$. $D$ is the dimension of visual representation output by $\boldsymbol{\varphi}_v$, so $\boldsymbol{h}_{c}$ and $\boldsymbol{o}_{c}$ can directly join in training. We denote the corresponding sets as $\mathcal{H}=\{\boldsymbol{h}_{1},\boldsymbol{h}_{2},...,\boldsymbol{h}_{C}\}$ and $\mathcal{O}=\{\boldsymbol{o}_{1},\boldsymbol{o}_{2},...,\boldsymbol{o}_{C}\}$, and $C$ is the number of ID classes.
To access updated $\mathcal{H}$ and $\mathcal{O}$ during training, we extract the embeddings of $\id$ before training and maintain an offline class-conditional queue with $N_c$ extracted embeddings from each ID class. In each iteration, we randomly replace a part of embeddings in each class queue with the same number of new embeddings.

\subsection{Bidirectional Prompt Customization}
To match diverse ID objects effectively, for the $c$-$th$ ID class, 
we concatenate the learnable class-specific prompts $\boldsymbol{W}_c\in \mathbb{R}^{K \times D}$ with the $D$-dimensional class word embeddings $\boldsymbol{e}_c$,  
where $K$ is the number of tokens. The textual embedding of the $c$-$th$ ID class is formed as $\boldsymbol{Z}_c=[\boldsymbol{W}_{c}^1,\boldsymbol{W}_{c}^2,...,\boldsymbol{W}_{c}^K,\boldsymbol{e}_c]$. Moreover, to enlarge the discrepancy between ID/OOD semantic, we expand $M$ learnable OOD (negative) prompts $\boldsymbol{\tilde{Z}} \in \mathbb{R}^{M \times K \times D}$ which to be aligned with samples in $\mathcal{O}$.
Then we obtain the ID and OOD textual representation denoted as $\boldsymbol{T^{id}}=\boldsymbol{\varphi}_t(\boldsymbol{Z}) \in \mathbb{R}^{C \times D}$ and $\boldsymbol{T^{ood}}=\boldsymbol{\varphi}_t(\boldsymbol{Z}) \in \mathbb{R}^{M \times D}$, where $\boldsymbol{\varphi}_t$ is the text encoder of CLIP. And the completed txetual representation is written as $\boldsymbol{T}=[\boldsymbol{T^{id}}, \boldsymbol{T^{ood}}] \in \mathbb{R}^{(C+M)\times D}$.

With the updated $\mathcal{H}$ and $\mathcal{O}$ in every iteration, we first consist $\mathcal{D}_{\text{few}}$ with 16 images for each ID class from $\id$ to limit the training cost. Then we form the visual embeddings set $\mathcal{V} = \{\boldsymbol{\varphi}_v(x_{i^{\prime}})|x_{i^{\prime}}\in\mathcal{D}_{\text{few}}\} \cup \mathcal{H}_s \cup \mathcal{O}$ for alignment with $\boldsymbol{T}$ through corss-entropy (CE) loss. Note that to maintain the class distribution in a batch of training data, we randomly select $S$ samples from $\mathcal{H}$ to constitute $\mathcal{H}_s$ and $S$ is half of batchsize  ($S<C$). For one embedding $ \boldsymbol{v} \in \mathbb{R}^{D}$ from $\mathcal{V}$, the prediction probability on the ID classes and the expanded OOD context is formed as:
\vspace{-0.3em}
\begin{equation}
p(\boldsymbol{y}|\boldsymbol{v})=\frac{\exp(\boldsymbol{s_v} / \tau)}{\sum_{k=1}^{C+M}\exp(s_v^k / \tau)} \in \mathbb{R}^{(C+M)}, 
\vspace{-0.3em}
\label{eq:prob_total}
\end{equation}
where $\tau$ is a temperature parameter learned by CLIP. $\boldsymbol{s_v}=[s_v^1,s_v^2,...,s_v^{C+M}]$ and $s_v^k=\cos(\boldsymbol{v},\boldsymbol{T}_k)$. In terms of ground truth labels in CE loss, we assign random labels from $\mathcal{Y_O} = \{C,C+1,...,C+M-1\}$ to samples in $\mathcal{O}$, while labels $\mathcal{Y_H}$ in $\mathcal{H}_s$ are the index of selected samples from $\mathcal{H}$ and $\mathcal{Y_H} \in \idlabel=\{0,1,...,C-1\}$. CE loss is formulated as: 
\vspace{-0.3em}
\begin{equation}
\mathcal{L}_{ce} = -\mathbb{E}_{(\boldsymbol{v},y)\sim \mathcal{V}}[\ell_{\mathrm{CE}}(p(\boldsymbol{y}|\boldsymbol{v}),y)].
\vspace{-0.2em}
\label{eq:celoss}
\end{equation}
Meanwhile, we compute the similarity of ID semantic-relevant region $\boldsymbol{h} \in \mathcal{H}$ on OOD context as:
\vspace{-0.3em}
\begin{equation}
p(\boldsymbol{y_{ood}}|\boldsymbol{h})=\frac{\exp(\boldsymbol{s_h} / \tau)}{\sum_{k=1}^{M}\exp(s_h^k / \tau)} \in \mathbb{R}^{M},
\vspace{-0.1em}
\label{eq:prob_out}
\end{equation}
where $\boldsymbol{s_h}=[s_h^1,s_h^2,...,s_h^M]$ and $s_h^k=\cos(\boldsymbol{h},\boldsymbol{T^{ood}}_k)$. Then we employ the OE loss \cite{hendrycks18oe} to unify the prediction of $\boldsymbol{h}$ on the extended negative context: 
\vspace{-0.3em}
\begin{equation}
\mathcal{L}_{uni} = -\mathbb{E}_{\boldsymbol{h}\sim \mathcal{H}}[\ell_{\mathrm{OE}}(p(\boldsymbol{y_{ood}}|\boldsymbol{h}))],
\vspace{-0.15em}
\label{eq:uniloss}
\end{equation}
where $\ell_{\mathrm{OE}}(p(\boldsymbol{y_{ood}}|\boldsymbol{h}))= \frac{\mathbf{1}_M}{M} \cdot \log(\boldsymbol{y_{ood}}|\boldsymbol{h}))$, and $\frac{\mathbf{1}_M}{M}$ is the uniform posterior distribution over all of $M$ OOD textual vectors.

To motivate $\boldsymbol{T^{id}}$ and $\boldsymbol{T^{ood}}$ to be further apart without a metric loss needing a margin hyper-parameter, we introduce a binary sigmoid loss formulated as:
\vspace{-0.3em}
\begin{equation}
\begin{aligned}
\mathcal{L}_{bin}&=\mathbb{E}_{\boldsymbol{h}\sim \mathcal{H}}\left[- \log({\rm S}(\mathop{\max}\limits_{k} \ \cos(\boldsymbol{h},\boldsymbol{T^{id}}_k))) \right]+\\
&\mathbb{E}_{\boldsymbol{h}\sim \mathcal{H}}\left[- \log(1-{\rm S}(\mathop{\max}\limits_{k} \ \cos(\boldsymbol{h},\boldsymbol{T^{ood}}_k))) \right],
\vspace{-0.5em}
\label{eq:binsloss}
\end{aligned}
\end{equation}
where ${\rm S}(\cdot)$ is the sigmoid function.

Our overall objective can be expressed as~\cref{eq:overall_loss} with the weights $\gamma$ and $\lambda$.
\vspace{-0.3em}
\begin{equation}\label{eq:overall_loss}
\mathcal{L} = \mathcal{L}_{ce} + \gamma \mathcal{L}_{uni} + \lambda \mathcal{L}_{bin}.
\end{equation}

\subsection{Test-time OOD Score}
During evaluation, we propose the D-energy score for OOD detection, which is the difference between ID-related energy and OOD-related energy:
\vspace{-0.5em}
\begin{equation}\label{e:Denergy}
{\rm D\mbox{-}energy}(xi)={\rm E}^{id}(x_i)-{\rm E}^{ood}(x_i).
\vspace{-0.3em}
\end{equation}
${\rm E}^{id}(x_i)$ and ${\rm E}^{ood}(x_i)$ are the energy functions written as:
\vspace{-0.5em}
\begin{equation}
\label{eq:idenergy}
{\rm E}^{id}(x_i)=T \cdot \log\sum_{k=1}^C \exp(s_{id}^k / T),
\end{equation}
\vspace{-0.5em}
\begin{equation}
\label{eq:oodenergy}
{\rm E}^{ood}(x_i)=T \cdot \log\sum_{k=1}^M \exp(s_{ood}^k / T),
\vspace{-0.3em}
\end{equation}
where $T$ is the temperature parameter and we apply its default value of 1. $s_{id}^k$ and $s_{ood}^k$ are the similarities of the test data with ID and OOD contexts, which are formed as:
\vspace{-0.3em}
\begin{equation}
\begin{aligned}
\label{eq:idsim}
s_{id}^k&=\cos(\boldsymbol{\varphi}_v(x_i),\boldsymbol{T^{id}}_k), k \in \{1,2,...,C\}, \\
s_{ood}^k&=\cos(\boldsymbol{\varphi}_v(x_i),\boldsymbol{T^{ood}}_k), k \in \{1,2,...,M\}. 
\vspace{-0.4em}
\end{aligned}
\end{equation}
ID data from $\idtest$ and $\csIDtest$ will obtain higher ${\rm E}^{id}(x_i)$ considering their larger similarity with $\boldsymbol{T^{id}}$ than OOD data, while the ${\rm E}^{ood}(x_i)$ of ID data will be lower because $\boldsymbol{T^{ood}}$ is pushed away from $\boldsymbol{T^{ood}}$ through $\mathcal{L}_{bin}$ in \cref{eq:binsloss}. Finally, ID data will be promoted to produce higher ID-ness on the D-energy score which subtracts ${\rm E}^{ood}(x_i)$ from ${\rm E}^{id}(x_i)$.

In F-OOD, $\nearoodtest$ is more complex to detect owning to its similar domain and style with $\id$ while $\faroodtest$ is more tractable considering its covariate shift. Benefit from our learned prompts with sensitivity to semantic shift and robustness under covariate shift, we introduce the extra MCM score \cite{mcm} to detect samples from $\nearoodtest$:
\vspace{-0.4em}
\begin{equation}
\begin{gathered}
\label{eq:mcm}
{\rm MCM}(xi)=\mathop{\max}\limits_{k} \ \cos(\boldsymbol{\varphi}_v(x_i),\boldsymbol{T^{id}}_k),  \\ 
k \in \{1,2,...,C\}.
\vspace{-0.2em}
\end{gathered}
\end{equation}

\section{Experiments}\label{sec:exp}

\begin{table*}[btp!]
\renewcommand{\arraystretch}{1.}
\renewcommand{\tabcolsep}{8.pt}
\centering
\small
\caption{\textbf{Comparison between previous methods and LSA on two F-OOD benchmarks.} In each row,  we report the averaged values on two Near-OOD datastes and three Far-OOD datasets, while the dataset-specific values are shown in Appendix. $\uparrow$/$\downarrow$ indicates higher/lower value is better, and the best results are in \textbf{bold}.}
\label{table:MainResults}
\begin{tabular}{c|cc|cc|cc|c}
\toprule 
                          \multicolumn{8}{c}{\textbf{Benchmark: ImageNet-200}}         \\  \cmidrule(lr){1-8}
                          & \multicolumn{2}{c|}{FPR@95~$\downarrow$}                                   & \multicolumn{2}{c|}{AUROC~$\uparrow$}                               & \multicolumn{2}{c|}{AUPR-IN~$\uparrow$}                        &                       \\ \cmidrule(lr){2-7}
\multirow{-2}{*}{Methods} & Near-OOD                      & Far-OOD                       & Near-OOD                      & Far-OOD                  & Near-OOD                 & Far-OOD                  & \multirow{-2}{*}{ACC~$\uparrow$} \\ \cmidrule(lr){1-8}
MCD \cite{mcd}                      & \cellcolor[HTML]{E6ECE3}89.15 & \cellcolor[HTML]{ffefe0}89.57 & \cellcolor[HTML]{E6ECE3}58.63 & \cellcolor[HTML]{ffefe0}65.07 & \cellcolor[HTML]{E6ECE3}73.08 & \cellcolor[HTML]{ffefe0}89.96   &     43.95                  \\
\small{NPOS \cite{npos2023iclr}+KNN \cite{sun2022knnood}}                  & \cellcolor[HTML]{E6ECE3}94.99 & \cellcolor[HTML]{ffefe0}65.04      & \cellcolor[HTML]{E6ECE3}51.19      & \cellcolor[HTML]{ffefe0}78.99 & \cellcolor[HTML]{E6ECE3}69.44 & \cellcolor[HTML]{ffefe0}93.24 &            /           \\
\small{PixMix \cite{hendrycks2021pixmix}+RMDS \cite{rmd21arxiv}}                          & \cellcolor[HTML]{E6ECE3}91.30 & \cellcolor[HTML]{ffefe0}93.60      & \cellcolor[HTML]{E6ECE3}61.91      & \cellcolor[HTML]{ffefe0}65.43 & \cellcolor[HTML]{E6ECE3}74.79 & \cellcolor[HTML]{ffefe0}90.35 &            49.36           \\
Gram \cite{gram20icml}                         & \cellcolor[HTML]{E6ECE3}91.02      & \cellcolor[HTML]{ffefe0}83.89      & \cellcolor[HTML]{E6ECE3}61.23      & \cellcolor[HTML]{ffefe0}65.26 & \cellcolor[HTML]{E6ECE3}75.16 & \cellcolor[HTML]{ffefe0}87.50 &   43.92   \\ \midrule
MCM \cite{mcm}                        & \cellcolor[HTML]{E6ECE3}87.24      & \cellcolor[HTML]{ffefe0}62.43      & \cellcolor[HTML]{E6ECE3}70.07      & \cellcolor[HTML]{ffefe0}88.99 & \cellcolor[HTML]{E6ECE3}79.58 & \cellcolor[HTML]{ffefe0}97.20 &         76.37              \\
CoOp \cite{coop}                         & \cellcolor[HTML]{E6ECE3}91.84      & \cellcolor[HTML]{ffefe0}67.60      & \cellcolor[HTML]{E6ECE3}63.69      & \cellcolor[HTML]{ffefe0}85.75 & \cellcolor[HTML]{E6ECE3}76.11 & \cellcolor[HTML]{ffefe0}95.92 &         73.41              \\
CoCoOp \cite{cocoop}                         & \cellcolor[HTML]{E6ECE3}89.07      & \cellcolor[HTML]{ffefe0}70.29      & \cellcolor[HTML]{E6ECE3}67.17      & \cellcolor[HTML]{ffefe0}88.42 & \cellcolor[HTML]{E6ECE3}77.37 & \cellcolor[HTML]{ffefe0}96.93 &         74.32              \\
MaPLe \cite{khattak2023maple}                         & \cellcolor[HTML]{E6ECE3}91.19      & \cellcolor[HTML]{ffefe0}72.72      & \cellcolor[HTML]{E6ECE3}67.28      & \cellcolor[HTML]{ffefe0}86.19 & \cellcolor[HTML]{E6ECE3}79.10 & \cellcolor[HTML]{ffefe0}96.27 &            73.17           \\
LoCoOp \cite{miyai2023locoop}                         & \cellcolor[HTML]{E6ECE3}81.18      & \cellcolor[HTML]{ffefe0}49.82      & \cellcolor[HTML]{E6ECE3}70.15      & \cellcolor[HTML]{ffefe0}89.10 & \cellcolor[HTML]{E6ECE3}79.16 & \cellcolor[HTML]{ffefe0}96.76 &      75.33                 \\ \midrule
\textbf{LSA(ours)}                        & \cellcolor[HTML]{E6ECE3}\textbf{52.73}      & \cellcolor[HTML]{ffefe0}\textbf{32.41}      & \cellcolor[HTML]{E6ECE3}\textbf{85.41}      & \cellcolor[HTML]{ffefe0}\textbf{90.97} & \cellcolor[HTML]{E6ECE3}\textbf{90.26} & \cellcolor[HTML]{ffefe0}\textbf{97.49} &       \textbf{76.54}         \\      

\bottomrule
\end{tabular} 

\begin{tabular}{c|cc|cc|cc|c}
\toprule
                          \multicolumn{8}{c}{\textbf{Benchmark: ImageNet-1K}}         \\  \cmidrule(lr){1-8}
                          & \multicolumn{2}{c|}{FPR@95~$\downarrow$}                                   & \multicolumn{2}{c|}{AUROC~$\uparrow$}                               & \multicolumn{2}{c|}{AUPR-IN~$\uparrow$}                        &                       \\ \cmidrule(lr){2-7}
\multirow{-2}{*}{Methods} & Near-OOD                      & Far-OOD                       & Near-OOD                      & Far-OOD                  & Near-OOD                 & Far-OOD                  & \multirow{-2}{*}{ACC~$\uparrow$} \\ \cmidrule(lr){1-8}
\small{DeepAug \cite{deepaug}+SHE \cite{she23iclr}}                    & \cellcolor[HTML]{E6ECE3}83.26 & \cellcolor[HTML]{ffefe0}68.70 & \cellcolor[HTML]{E6ECE3}68.27 & \cellcolor[HTML]{ffefe0}78.85 & \cellcolor[HTML]{E6ECE3}89.11 & \cellcolor[HTML]{ffefe0}96.49 & 57.82            \\
\footnotesize{StyAug \cite{styaug}+GradNorm \cite{gradnorm}}     & \cellcolor[HTML]{E6ECE3}87.14 & \cellcolor[HTML]{ffefe0}58.82      & \cellcolor[HTML]{E6ECE3}65.27      & \cellcolor[HTML]{ffefe0}81.62 & \cellcolor[HTML]{E6ECE3}87.99 & \cellcolor[HTML]{ffefe0}96.81 &   55.44       \\
\small{AugMix \cite{hendrycks2019augmix}+SHE \cite{she23iclr}}     & \cellcolor[HTML]{E6ECE3}84.45 & \cellcolor[HTML]{ffefe0}60.26      & \cellcolor[HTML]{E6ECE3}69.66      & \cellcolor[HTML]{ffefe0}83.06 & \cellcolor[HTML]{E6ECE3}89.29 & \cellcolor[HTML]{ffefe0}97.24 &   57.46    \\
ASH \cite{ash}                          & \cellcolor[HTML]{E6ECE3}93.27      & \cellcolor[HTML]{ffefe0}59.56      & \cellcolor[HTML]{E6ECE3}60.52      & \cellcolor[HTML]{ffefe0}86.75 & \cellcolor[HTML]{E6ECE3}85.41 & \cellcolor[HTML]{ffefe0}97.15 &            54.35           \\ \midrule
MCM \cite{mcm}   & \cellcolor[HTML]{E6ECE3}94.74      & \cellcolor[HTML]{ffefe0}77.47      & \cellcolor[HTML]{E6ECE3}58.11      & \cellcolor[HTML]{ffefe0}82.56 & \cellcolor[HTML]{E6ECE3}83.87 & \cellcolor[HTML]{ffefe0}97.84 &            58.52           \\
CoOp \cite{coop}      & \cellcolor[HTML]{E6ECE3}95.82      & \cellcolor[HTML]{ffefe0}80.31      & \cellcolor[HTML]{E6ECE3}56.36      & \cellcolor[HTML]{ffefe0}82.41 & \cellcolor[HTML]{E6ECE3}83.76 & \cellcolor[HTML]{ffefe0}97.64 &           61.22            \\
CoCoOp \cite{cocoop}    & \cellcolor[HTML]{E6ECE3}96.23      & \cellcolor[HTML]{ffefe0}79.59      & \cellcolor[HTML]{E6ECE3}53.11      & \cellcolor[HTML]{ffefe0}74.48 & \cellcolor[HTML]{E6ECE3}82.81 & \cellcolor[HTML]{ffefe0}94.87 &             60.95          \\
MaPLe \cite{khattak2023maple}     & \cellcolor[HTML]{E6ECE3}95.23      & \cellcolor[HTML]{ffefe0}75.84      & \cellcolor[HTML]{E6ECE3}57.38      & \cellcolor[HTML]{ffefe0}83.70 & \cellcolor[HTML]{E6ECE3}84.38 & \cellcolor[HTML]{ffefe0}97.84 &           \textbf{62.15}            \\
LoCoOp \cite{miyai2023locoop}    & \cellcolor[HTML]{E6ECE3}90.91      & \cellcolor[HTML]{ffefe0}54.33      & \cellcolor[HTML]{E6ECE3}59.34      & \cellcolor[HTML]{ffefe0}84.02 & \cellcolor[HTML]{E6ECE3}84.66 & \cellcolor[HTML]{ffefe0}97.92 &           59.83            \\ \midrule
\textbf{LSA(ours)}                          & \cellcolor[HTML]{E6ECE3}\textbf{70.56}      & \cellcolor[HTML]{ffefe0}\textbf{48.06}      & \cellcolor[HTML]{E6ECE3}\textbf{78.22}      & \cellcolor[HTML]{ffefe0}\textbf{86.85} & \cellcolor[HTML]{E6ECE3}\textbf{93.05} & \cellcolor[HTML]{ffefe0}\textbf{98.10} &       61.74         \\      

\bottomrule
\end{tabular}
\vspace{3pt}
\end{table*}

\subsection{Benchmarks and Compared Methods}
The full-spectrum out-of-distribution (F-OOD) detection benchmarks are first proposed in SEM \cite{yang2023full}, but the benchmarks in SEM are limited by the data scale and are unavailable in the \href{https://github.com/Jingkang50/OpenOOD} {official repository}. Hence, `F-OOD' in this work refers to the full-spectrum benchmarks in OpenOODv1.5 \cite{zhang2023openood} by default.
F-OOD benchmarks include the large-scale ImageNet-200 benchmark and ImageNet-1K benchmark, which regard ImageNet-200 and ImageNet-1K \cite{deng2009imagenet} as ID training data $\id$, respectively. They both incorporate ImageNet-C \cite{imagenetc} with image corruptions, ImageNet-R \cite{deepaug} with style changes and ImageNet-V2 \cite{imagenetv2} with resampling bias as $\csIDtest$. In terms of $\nearoodtest$, the two benchmarks include SSB-hard \cite{SSB} and NINCO \cite{ninco}. Additionally, they consider iNaturalist \cite{inaturalist}, Textures \cite{texture} and OpenImage-O \cite{haoqi2022vim} as $\faroodtest$.
 
We first select methods, which achieve the best result on FPR@95, AUROC and AUPR-IN of Near-OOD and Far-OOD according to the results released by OpenOODv1.5 in \href{https://docs.google.com/spreadshe ets/d/1mTFrO-_STYBRcNMMEmHQrFPQzeg6S8Z2vRA8jawTwBw/edit?usp=sharing}{full results}, as compared methods. Then three CLIP propmt tuning methods \textit{e.g.,} CoOp \cite{coop}, CoCoOp \cite{cocoop} and MaPLe \cite{khattak2023maple} are included. In the area of OOD detection, the CLIP zero-shot method MCM \cite{mcm} and prompt tuning method LoCoOp \cite{miyai2023locoop} are also chosen as compared methods.

\subsection{Results on Full-Spectrum Benchmarks} 
We compare the results of our proposed approach with the compared methods in \cref{table:MainResults}. All experiments use CLIP based on ViT-B/16. We only report the average metric values on corresponding OOD datasets for each benchmark limited by space. Results show that our proposed LSA consistently obtains the best results across all OOD detection metrics. It’s worth noting that, in the troublesome Near-OOD of the two benchmarks, LSA significantly outperforms LoCoOp by $28.45\%$/$20.35\%$ on FRP@95 and $15.26\%$/$18.88\%$ on AUROC. In terms of the OOD generalization metric, \textit{i.e.}, the classification accuracy (ACC) of $\idtest$ and $\csIDtest$, LSA delivers the second best result and is slightly inferior to MaPLe on ImageNet-1K benchmark. We attribute this phenomenon to the `spurious OOD' \cite{spurious}, which indicates that the background frequently co-existing with ID objects but without ID feature can benefit classification accuracy, \textit{e.g.}, deep models may rely on the river or sea to recognize `ship' instead of learning the semantic feature of ships. Therefore, our method which highlights the ID semantic-relevant regions, can compromise in classification performance.

\begin{table}[btp!]
\centering
\setlength{\tabcolsep}{0.8mm}
{
\small
\caption{\textbf{Ablation study of loss functions on ImageNet-200 benchmark.} Our completed method combines $\mathcal{L}_{ce}$, $\mathcal{L}_{uni}$ and $\mathcal{L}_{bin}$. The best results are in \textbf{bold}.}
\label{T:lossAb}
\begin{tabular}{ccc|cc|cc}
\toprule 
& & & \multicolumn{2}{c|}{FPR@95~$\downarrow$}         & \multicolumn{2}{c}{AUROC~$\uparrow$}    \\      \cmidrule(lr){4-7}
\multirow{-2}{*}{$\mathcal{L}_{ce}$} & \multirow{-2}{*}{$\mathcal{L}_{uni}$} & \multirow{-2}{*}{$\mathcal{L}_{bin}$}  & \small{Near-OOD}   & \small{Far-OOD}    & \small{Near-OOD}   & \small{Far-OOD}    \\   \cmidrule(lr){1-7}
$\checkmark$ & & &   \cellcolor[HTML]{E6ECE3}68.94 & \cellcolor[HTML]{ffefe0}55.03      & \cellcolor[HTML]{E6ECE3}77.29     & \cellcolor[HTML]{ffefe0}88.44 \\
$\checkmark$ & $\checkmark$ & &  \cellcolor[HTML]{E6ECE3}60.52 & \cellcolor[HTML]{ffefe0}40.63     & \cellcolor[HTML]{E6ECE3}83.86      & \cellcolor[HTML]{ffefe0}89.39  \\
$\checkmark$ & & $\checkmark$ &  \cellcolor[HTML]{E6ECE3}59.35 & \cellcolor[HTML]{ffefe0}38.96     & \cellcolor[HTML]{E6ECE3}82.27     & \cellcolor[HTML]{ffefe0}89.70 \\
$\checkmark$ & $\checkmark$ & $\checkmark$ &  \cellcolor[HTML]{E6ECE3}\textbf{52.73} & \cellcolor[HTML]{ffefe0}\textbf{32.41}      & \cellcolor[HTML]{E6ECE3}\textbf{85.41}      & \cellcolor[HTML]{ffefe0}\textbf{90.97} \\
\bottomrule
\end{tabular}}
\vspace{-5pt}
\end{table}

\subsection{Ablation Study and Qualitative Analysis}

\textbf{Effectiveness of losses and ID semantic-relevant regions $\boldsymbol{h}$.} We first analyze the effect of each loss in our proposed LSA in \cref{T:lossAb}. $\mathcal{L}_{ce}$ enhances the alignment of ID semantic-relevant regions $\boldsymbol{h}$ with $\boldsymbol{T^{id}}$ and initially learns OOD prompts with the OOD regularization regions $\boldsymbol{o}$, achieving better results on Near-OOD than LoCoOp. $\mathcal{L}_{uni}$ which flattens the similarity of $\boldsymbol{h}$ with $\boldsymbol{T^{ood}}$ and $\mathcal{L}_{bin}$ that pushes $\boldsymbol{T^{id}}$ and $\boldsymbol{T^{ood}}$ apart from each other, both further enlarges the advantages based on $\mathcal{L}_{ce}$. Finally, our complete method combining the three losses boosts the performance to the best.

Then we show the effect of the ID semantic-relevant regions $\boldsymbol{h}$ sampled from Gaussians in \cref{T:CenAb}. Exp$\#5$ is the version of LSA.
In general, $\boldsymbol{h}$ will enhance the performance on Near-OOD $\nearoodtest$ which can't be easily detected considering its homologous domain with $\id$, while covariance information from global images can improve the detection to Far-OOD $\faroodtest$.
Exp$\#1$ which exploits no $\boldsymbol{h}$, performs well on $\faroodtest$ while failing on $\nearoodtest$ compared to Exp$\#5$. The comparison between Exp$\#2$ \mbox{--} $\#4$ and Exp$\#1$ shows that, the performance exhibits varying degrees of improvement on $\nearoodtest$ when $\boldsymbol{h}$ is involved into training. Exp$\#5$ achieves the best results on $\nearoodtest$ utilizing $\boldsymbol{h}$ in all three losses. Although LSA is a bit inferior on AUROC of $\faroodtest$ compared to Exp$\#1$, the result of this metric outperforms all compared methods as presented in \cref{table:MainResults}. Moreover, we pay more attention to the performance on $\nearoodtest$ considering the covariate shift in $\faroodtest$ is a `shortcut' for OOD detection.

\begin{table}[btp!]
\centering
\setlength{\tabcolsep}{1.3mm}
{
\small
\caption{\textbf{The effectiveness of ID semantic-relevant regions $\boldsymbol{h}$.} $\mathcal{L}_{ce\text{-}G}$ means excluding $\boldsymbol{h}$ from $\mathcal{H}_s$ in $\mathcal{L}_{ce}$. $\mathcal{L}_{uni\text{-}G}$ and $\mathcal{L}_{bin\text{-}G}$ mean replacing $\boldsymbol{h} \in \mathcal{H}$ with the same number of global image embeddings in the corresponding loss. LSA combines $\mathcal{L}_{ce}$, $\mathcal{L}_{uni}$ and $\mathcal{L}_{bin}$. `Near'/`Far' means `Near-OOD'/`Far-OOD'. We refer to each experiment by its index for brevity. The best results are in \textbf{bold} and the second best are \underline{underlined}.}
\label{T:CenAb}
\begin{tabular}{c|c|cc|cc}
\toprule 
& & \multicolumn{2}{c|}{FPR@95~$\downarrow$}         & \multicolumn{2}{c}{AUROC~$\uparrow$}    \\      \cmidrule(lr){3-6}
& \multirow{-2}{*}{\normalsize{Strategy}}  & \small{Near}   & \small{Far}    & \small{Near}   & \small{Far}    \\   \cmidrule(lr){1-6}
1: & $\mathcal{L}_{ce\text{-}G}$+$\mathcal{L}_{uni\text{-}G}$+$\mathcal{L}_{bin\text{-}G}$ &          \cellcolor[HTML]{E6ECE3}58.15 & \cellcolor[HTML]{ffefe0}\underline{33.12} & \cellcolor[HTML]{E6ECE3}82.23 & \cellcolor[HTML]{ffefe0}\textbf{91.50}   \\
2: & $\mathcal{L}_{ce}$+$\mathcal{L}_{uni\text{-}G}$+$\mathcal{L}_{bin\text{-}G}$ &          \cellcolor[HTML]{E6ECE3}56.92 & \cellcolor[HTML]{ffefe0}33.93 & \cellcolor[HTML]{E6ECE3}82.50 & \cellcolor[HTML]{ffefe0}90.13 \\
3: & $\mathcal{L}_{ce}$+$\mathcal{L}_{uni}$+$\mathcal{L}_{bin\text{-}G}$&   \cellcolor[HTML]{E6ECE3}56.03 & \cellcolor[HTML]{ffefe0}38.83      & \cellcolor[HTML]{E6ECE3}83.59     & \cellcolor[HTML]{ffefe0}89.29 \\
4: & $\mathcal{L}_{ce}$+$\mathcal{L}_{uni\text{-}G}$+$\mathcal{L}_{bin}$&  \cellcolor[HTML]{E6ECE3}55.07 & \cellcolor[HTML]{ffefe0}36.95     & \cellcolor[HTML]{E6ECE3}84.06      & \cellcolor[HTML]{ffefe0}89.46  \\
5: & \textbf{LSA(ours)} &  \cellcolor[HTML]{E6ECE3}\textbf{52.73} & \cellcolor[HTML]{ffefe0}\textbf{32.41}      & \cellcolor[HTML]{E6ECE3}\textbf{85.41}      & \cellcolor[HTML]{ffefe0}\underline{90.97} \\
\bottomrule
\end{tabular}}
\vspace{-5pt}
\end{table}

\textbf{Effect of MCM scores in different methods.} MCM score formed as \cref{eq:mcm} is the maximum similarity of data with ID text embedding $\boldsymbol{T^{id}}$ and can reflect the ability of learned ID context to capture semantics. We observe employing the additional MCM score in LSA boosts the performance on $\nearoodtest$ while encountering a decline on $\faroodtest$. This phenomenon occurs because the $\boldsymbol{T^{id}}$ in LSA is brought close to the semantic high-likelihood regions during optimization and may not take into account the non-semantic covariance information which improves the recognition of $\faroodtest$. Moreover, we analyze the effect of MCM score in LoCoOp, which is completely opposite to that of LSA as demonstrated in \cref{T:MCM}. This comparison indicates that LSA more efficiently obtains the prompts with sensitivity to semantic shift and robustness under covariate shift, and we can flexibly select OOD score at test times according to the difficulty of OOD. It notes that our method consistently outperforms LoCoOp irrespective of whether the MCM score are used.

\begin{table}[btp!]
\centering
\setlength{\tabcolsep}{1.1mm}
{
\small
\caption{\textbf{The effect of MCM scores in different methods.} MCM score is formulated in \cref{eq:mcm}. $\vartriangle$ denotes the subtraction results between with/without MCM score during evaluation. `Near'/`Far' means `Near-OOD'/`Far-OOD'. {\color{blue} Blue} and {\color{red} red} mean decrease and increase, respectively.}
\label{T:MCM}
\begin{tabular}{c|cc|cc|cc}
\toprule 
& \multicolumn{2}{c|}{\footnotesize{FPR@95~$\downarrow$}}    & \multicolumn{2}{c|}{\footnotesize{AUROC~$\uparrow$}}   & \multicolumn{2}{c}{\footnotesize{AUPR-IN~$\uparrow$}}    \\      \cmidrule(lr){2-7}
\multirow{-2}{*}{}  & \small{Near}   & \small{Far}    & \small{Near}   & \small{Far}  & \small{Near}   & \small{Far}   \\   \cmidrule(lr){1-7}
\footnotesize{LoCoOp w/o MCM} &   \cellcolor[HTML]{E6ECE3}\textbf{76.20} & \cellcolor[HTML]{ffefe0}66.79 & \cellcolor[HTML]{E6ECE3}\textbf{71.93} & \cellcolor[HTML]{ffefe0}79.57 & \cellcolor[HTML]{E6ECE3}\textbf{80.21} & \cellcolor[HTML]{ffefe0}93.63 \\
 \footnotesize{LoCoOp w/ MCM} & \cellcolor[HTML]{E6ECE3}81.18 & \cellcolor[HTML]{ffefe0}\textbf{49.82}      & \cellcolor[HTML]{E6ECE3}70.15     & \cellcolor[HTML]{ffefe0}\textbf{89.10}  & \cellcolor[HTML]{E6ECE3}79.16 & \cellcolor[HTML]{ffefe0}\textbf{96.76} \\
 \normalsize{\textbf{$\vartriangle$}}
 &   \cellcolor[HTML]{E6ECE3}{\color{red}+4.98}  & \cellcolor[HTML]{ffefe0}{\color{blue}-16.97}      & \cellcolor[HTML]{E6ECE3}{\color{blue}-1.78}     & \cellcolor[HTML]{ffefe0}{\color{red}+9.53}  & \cellcolor[HTML]{E6ECE3}{\color{blue}-1.05} & \cellcolor[HTML]{ffefe0}{\color{red}+3.13} \\
  \cmidrule(lr){1-7}
LSA w/o MCM &  \cellcolor[HTML]{E6ECE3}55.62 & \cellcolor[HTML]{ffefe0}\textbf{32.41}     & \cellcolor[HTML]{E6ECE3}82.35      & \cellcolor[HTML]{ffefe0}\textbf{90.97}   & \cellcolor[HTML]{E6ECE3}86.39 & \cellcolor[HTML]{ffefe0}\textbf{97.49} \\
LSA w/ MCM &  \cellcolor[HTML]{E6ECE3}\textbf{52.73} & \cellcolor[HTML]{ffefe0}42.14      & \cellcolor[HTML]{E6ECE3}\textbf{85.41}      & \cellcolor[HTML]{ffefe0}88.54 & \cellcolor[HTML]{E6ECE3}\textbf{90.26} & \cellcolor[HTML]{ffefe0}97.02 \\
  \normalsize{\textbf{$\vartriangle$}}  &  \cellcolor[HTML]{E6ECE3}{\color{blue}-2.89} & \cellcolor[HTML]{ffefe0}{\color{red}+9.73}      & \cellcolor[HTML]{E6ECE3}{\color{red}+3.06}      & \cellcolor[HTML]{ffefe0}{\color{blue}-2.43}   & \cellcolor[HTML]{E6ECE3}{\color{red} +3.87} & \cellcolor[HTML]{ffefe0}{\color{blue}-0.47} \\
\bottomrule
\end{tabular}}
\vspace{-5pt}
\end{table}

\begin{figure}[t]
\small
\centering
\includegraphics[width=1.00\linewidth]{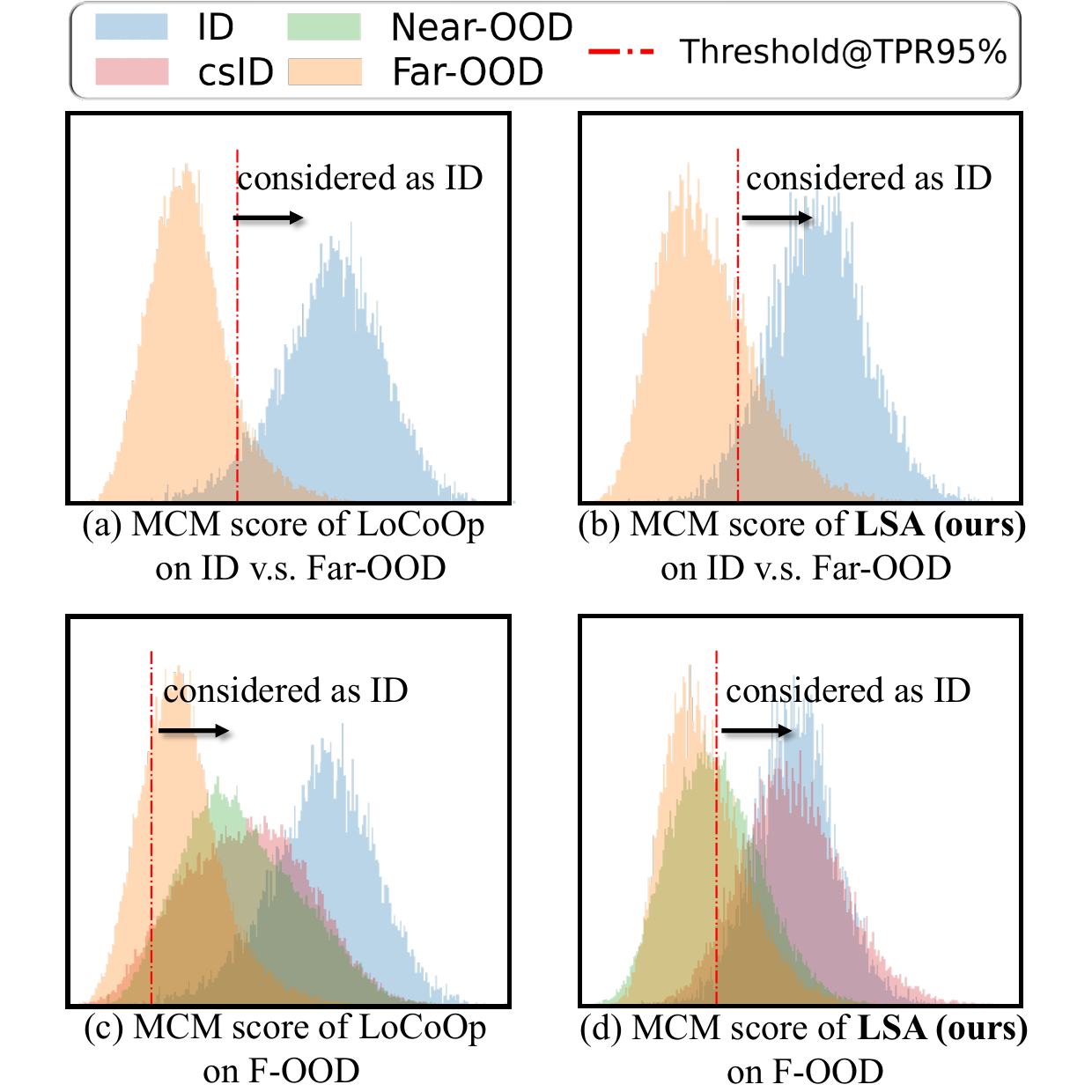}
\vspace{-16pt}
\caption{\textbf{Comparative distributions of MCM score learned by LoCoOp and LSA on F-OOD.} The comparison between (a) and (c) shows that when encountering csID and Near-OOD data, LoCoOp is confused in their distributions on MCM score, where csID samples severely overlap with Near-OOD while are far from ID, and the two groups of OOD also distributes differently from each other. However, as shown in (d), csID and ID samples share a similar distribution, and Near-OOD maintains consistency with Far-OOD on the MCM score learned by our proposed LSA.} 
\vspace{-3pt}
\label{fig:MCM2}
\end{figure}

We also present the distribution of data from F-OOD on the MCM score learned by LoCoOp and LSA in \cref{fig:MCM2}. When only facing $\idtest$ and $\faroodtest$, which clearly differ in both semantics and covariance, the MCM score from LoCoOp more significantly distinguishes between the data groups, and fewer samples from $\faroodtest$ are misclassified into ID as demonstrated by the comparison between \cref{fig:MCM2}\textcolor{red}{(a)} and \cref{fig:MCM2}\textcolor{red}{(b)}. However, when F-OOD introduces $\csIDtest$ and $\nearoodtest$, LoCoOp suffers from the severe overlap between data with different semantics on its MCM score as shown in \cref{fig:MCM2}\textcolor{red}{(c)}. Gratifyingly, the MCM score based on our learned $\boldsymbol{T^{id}}$ captures the consistency of semantics as \cref{fig:MCM2}\textcolor{red}{(d)} shows, two groups of ID data maintain a similar distribution, and the same goes for $\nearoodtest$ and $\faroodtest$.

\textbf{Influence of OOD scores.} To exclude the possibility that the performance of our proposed LSA only gains from the OOD score we applied, we compare the performance of LoCoOp \cite{miyai2023locoop} with LSA across several OOD scores, including MCM, MCM-GL and energy. As shown in \cref{T:score}, LSA consistently outperforms LoCoOp no matter which OOD score is chosen, even with the MCM-GL score boosting the OOD detection performance of LoCoOp. 
Moreover, our D-energy score in \cref{e:Denergy} which utilizes the $\boldsymbol{T^{ood}}$ away from ID semantic, further improves the OOD detection performance based on the energy score.


\begin{table}[btp!]
\centering
\setlength{\tabcolsep}{0.2mm}
{
\small
\caption{\textbf{Performance comparison between LoCoOp and LSA across different OOD scores.} `MCM-GL' is the OOD score applied in LoCoOp \cite{miyai2023locoop}, which is the sum of the MCM score and the maximum similarity of the patch with $\boldsymbol{T^{id}}$. `Energy' and `D-energy' are shown in \cref{eq:idenergy} and \cref{e:Denergy}, respectively.}
\label{T:score}
\begin{tabular}{cc|cc|cc}
\toprule 
& & \multicolumn{2}{c|}{FPR@95~$\downarrow$}         & \multicolumn{2}{c}{AUROC~$\uparrow$}    \\      \cmidrule(lr){3-6}
\multirow{-2}{*}{Score} & \multirow{-2}{*}{Methods} & \small{Near-OOD}   & \small{Far-OOD}    & \small{Near-OOD}   & \small{Far-OOD}    \\   \cmidrule(lr){1-6}
\multirow{2}{*}{MCM}  &  \footnotesize{LoCoOp \cite{miyai2023locoop}} &      \cellcolor[HTML]{E6ECE3}91.40 & \cellcolor[HTML]{ffefe0}69.26 & \cellcolor[HTML]{E6ECE3}65.28 & \cellcolor[HTML]{ffefe0}86.12  \\
  & $\footnotesize{\textbf{LSA(ours)}}$  &   \cellcolor[HTML]{E6ECE3}56.61 & \cellcolor[HTML]{ffefe0}52.78      & \cellcolor[HTML]{E6ECE3}81.53     & \cellcolor[HTML]{ffefe0}84.44 \\   \cmidrule(lr){1-6}
 \multirow{2}{*}{MCM-GL} &  \footnotesize{LoCoOp \cite{miyai2023locoop}} &      \cellcolor[HTML]{E6ECE3}81.18 & \cellcolor[HTML]{ffefe0}49.82 & \cellcolor[HTML]{E6ECE3}70.15 & \cellcolor[HTML]{ffefe0}89.10  \\
 &  $\footnotesize{\textbf{LSA(ours)}}$ &  \cellcolor[HTML]{E6ECE3}57.31 & \cellcolor[HTML]{ffefe0}40.75     & \cellcolor[HTML]{E6ECE3}80.24      & \cellcolor[HTML]{ffefe0}89.74  \\   \cmidrule(lr){1-6}
 \multirow{2}{*}{Energy}  &  \footnotesize{LoCoOp \cite{miyai2023locoop}} &  \cellcolor[HTML]{E6ECE3}88.46 & \cellcolor[HTML]{ffefe0}90.74 & \cellcolor[HTML]{E6ECE3}63.03 & \cellcolor[HTML]{ffefe0}72.51  \\
   & $\footnotesize{\textbf{LSA(ours)}}$  &   \cellcolor[HTML]{E6ECE3}56.85 & \cellcolor[HTML]{ffefe0}53.41     & \cellcolor[HTML]{E6ECE3}81.81     & \cellcolor[HTML]{ffefe0}82.32 \\   \cmidrule(lr){1-6}
 D-energy & $\footnotesize{\textbf{LSA(ours)}}$ &  \cellcolor[HTML]{E6ECE3}\textbf{55.62} & \cellcolor[HTML]{ffefe0}\textbf{32.41}      & \cellcolor[HTML]{E6ECE3}\textbf{82.35}      & \cellcolor[HTML]{ffefe0}\textbf{90.97} \\
\bottomrule
\end{tabular}}
\vspace{2pt}
\end{table}

\begin{figure}[t]
\centering
\includegraphics[width=0.98\linewidth]{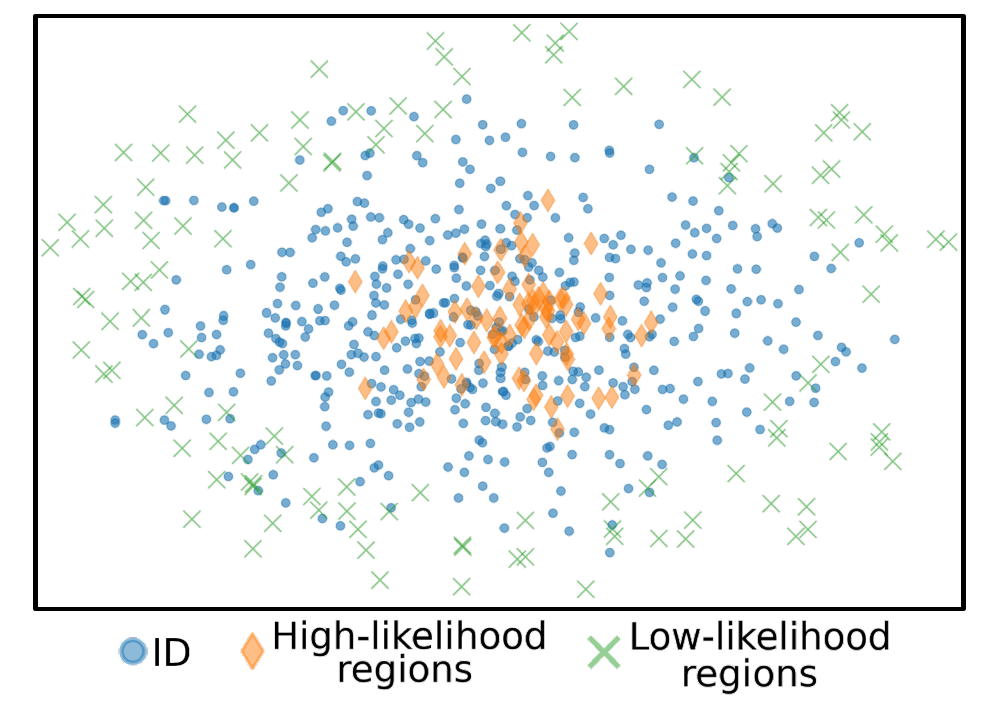}
\vspace{-10pt}
\caption{\textbf{UMAP visualization of the sampled regions from Gaussians.} We visualize the distribution of `goose' ID data, high-likelihood and low-likelihood regions from `goose' Gaussians. High-likelihood regions are more compact than global ID feature and more consistent with the class center, while low-likelihood regions are concentrated in the boundary far from the class center.}
\vspace{-10pt}
\label{fig:UMAP}
\end{figure}

\begin{figure}[t]
\small
\centering
\begin{overpic}[width=1\linewidth]{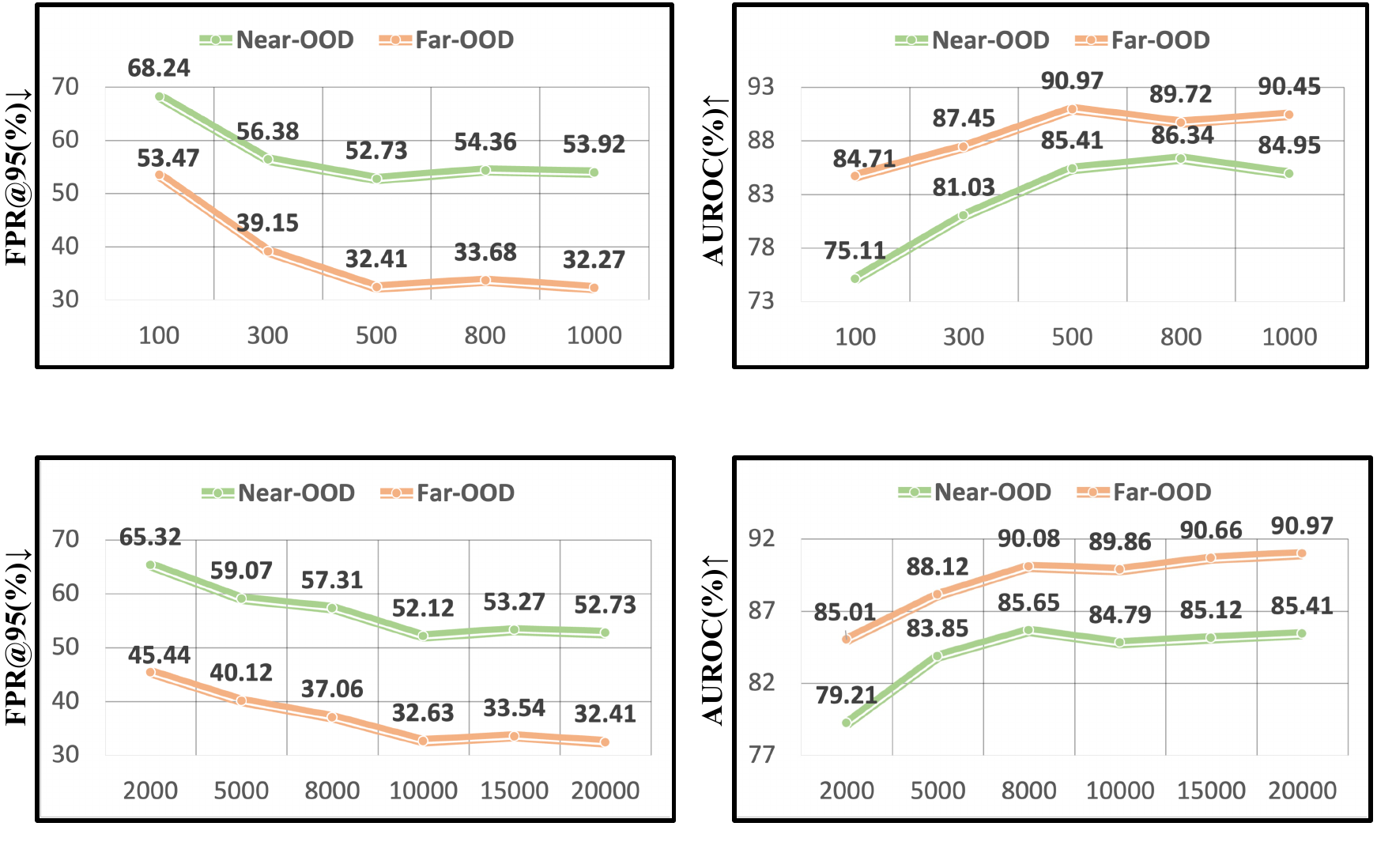}
        \put(25, 33){(a) Comparison between different $N_c$}
        \put(25, 0){(b) Comparison between different $N$}
\end{overpic}
\vspace{-5pt}
\caption{\textbf{Analysis on important hyper-parameters.} We analyze the effect of the size of each class queue $N_c$ and the number of sampled data $N$ from each class Gaussians. $\uparrow$/$\downarrow$ indicates higher/lower value is better. (a) and (b) both prove the the proposed method can obtain a stably good result when $N_c\ge500$ and $N\ge10000$, showing the robustness of our method.} 
\vspace{-10pt}
\label{fig:gaussianhyper}
\end{figure}

\textbf{Visualization of the sampled regions from Gaussians.} As illustrated in the visualization using UMAP \cite{mcinnes2018umap} in \cref{fig:UMAP}, high-likelihood regions sampled from Gaussians are more compact and more proximate to the class feature center compared to global ID images, indicating that they are more suitable for alignment with ID class context. Moreover, low-likelihood regions can be exploited as OOD regularization since they are located far from class center.

\textbf{Analysis on important hyper-parameters.} Here we analyze the impact of the hyper-parameters in the offline Gaussians sampling strategy, including the size of each class queue $N_c$ and the number of sampled data $N$ from each class Gaussians. In general, a larger $N_c$ used to calculate the statistics is more beneficial to estimate the precise class-conditional Gaussian distribution. \cref{fig:gaussianhyper}{\textcolor{red}{(a)}}  shows the performance fluctuation is slight when $N_c\ge500$, and we choose $N_c=500$. Although every class contains about 1,200 training samples in ImageNet-1k and ImageNet-200, the offline embedding queue extracted by the pre-trained encoder does not need to be optimized and the embeddings from a part of samples are sufficient to model an accurate Gaussian distribution.

Moreover, we sample regions with the highest/lowest probability density from $N$ data in the Gaussians, and the performance is not sensitive to $N$ when $N\ge10000$ as shown in \cref{fig:gaussianhyper}{\textcolor{red}{(b)}}. $N$ is not hard to choose considering the $3\sigma$ principle, which suggests about $99.7\%$ of data points in Gaussians fall within three standard deviations ($\sigma$) of the mean ($\mu$). As a result, we can precisely sample data located around $\mu$ and $\mu\pm3\sigma$ when $N$ is large enough. More analysis experiences about other hyper-parameters are shown in Appendix.
\vspace{0.2em}
\section{Conclusion}\label{sec:conclusion}
\vspace{0.2em}
In this work, we focus on addressing the realistic F-OOD detection task, where models are required to precisely perceive ID semantic when disturbed by both semantic and covariate shifts. We reveal that high-likelihood regions concentrated around the ID or csID class centers maintain a solid semantic consistency, which contributes to mitigating the interference from covariate shift.
We propose the LSA framework that efficiently captures semantically high-likelihood visual regions and adaptively optimizes the prompt-based decision boundary. Experimental results indicate that LSA effectively promotes the image-text alignment to focus on semantic shift,
achieving superior performance than existing OOD detection and prompt learning methods on F-OOD benchmarks.

\newpage

{
\small
\bibliographystyle{ieeenat_fullname}
\bibliography{ref.bib}
}

\clearpage
\appendix
\renewcommand\thefigure{A\arabic{figure}}
\renewcommand\thetable{A\arabic{table}}
\renewcommand\theequation{A\arabic{equation}}
\setcounter{equation}{0}
\setcounter{table}{0}
\setcounter{figure}{0}
\section*{Appendix}

\section{Experiment Details}
Following existing works \cite{zeroMCMlocal,miyai2023locoop,coop}, the CLIP based on ViT-B/16 \cite{radford2021learning} is employed for all experiments, which is trained by an SGD optimizer with a weight decay of $0.0005$ and a momentum of $0.9$. We use the cosine annealing learning rate starting at $0.004$, taking totally $100$ epochs. The dataloader is prepared with a batch-size of $64$ for the ID training set $\id$. For the size of each class queue $N_c$ and the number of sampled data $N$ from each class Gaussians, we use $500$ and $20000$, respectively. The number of the expanded OOD contexts $M$ is set as $15$ and we apply the number of tokens $K=3$ for both learnable ID and OOD contexts. The analysis experiences about $M$ and $K$ are shown in \cref{sec:hyperparameters}. The learnable ID and OOD contexts are both initialized randomly.

For the training objective of our proposed LSA is denoted as:
\begin{equation}\label{e:overall_loss}
\mathcal{L} = \mathcal{L}_{ce} + \gamma \mathcal{L}_{uni} + \lambda \mathcal{L}_{bin},
\end{equation}
where we set $\gamma=0.5$ and $\lambda=0.1$ for all experiments.

\section{Visualization of ID and csID Features}
\begin{figure}[t]
\centering
\includegraphics[width=0.98\linewidth]{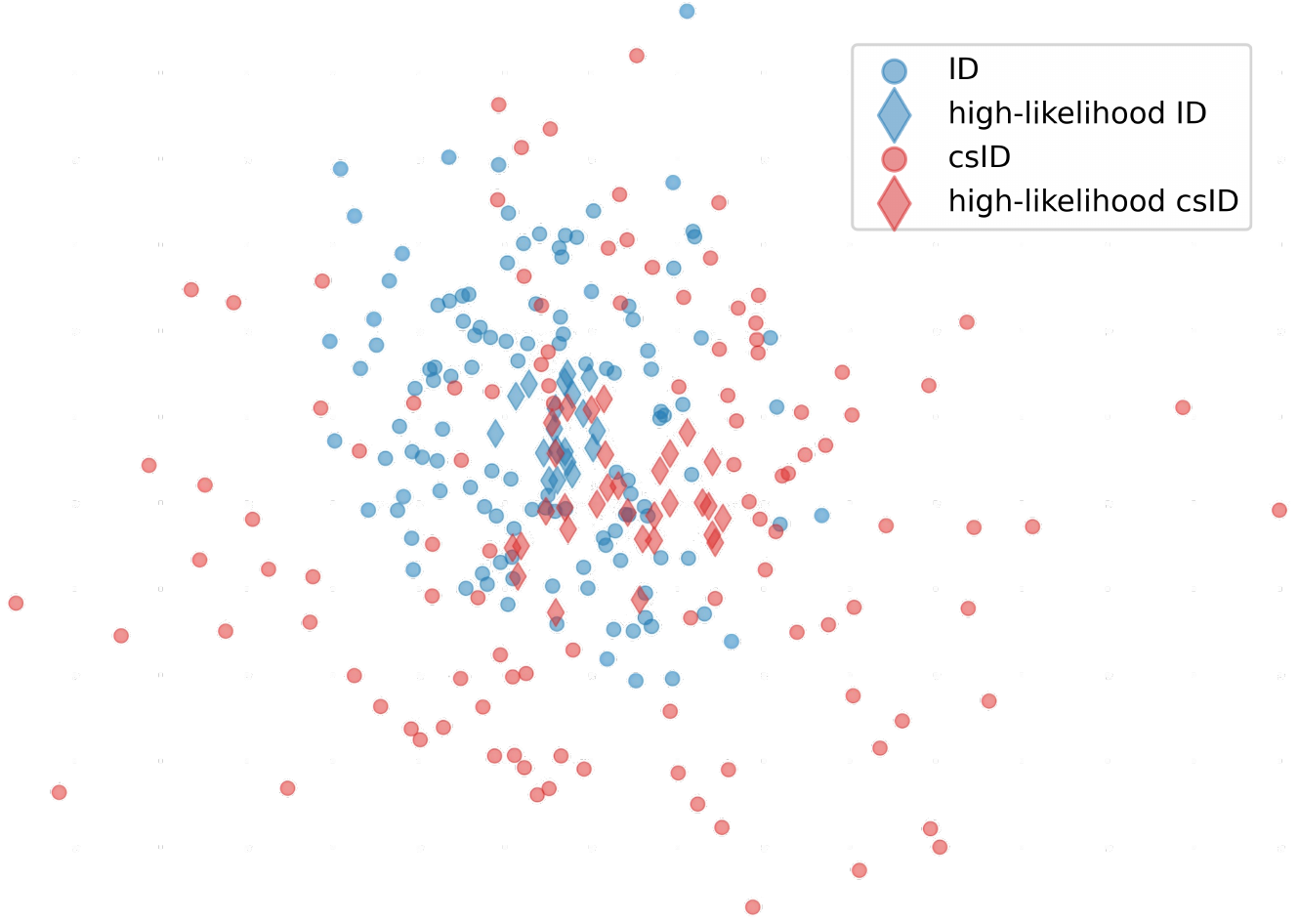}
\vspace{-5pt}
\caption{\textbf{Visualization of the ID and csID features.} We visualize the `goose' features from ID and csID data with UMAP \cite{mcinnes2018umap}. And we denote the 20 high-likelihood ID/csID features closest to the corresponding class center as diamonds. High-likelihood features from ID and csID are more compact and consistent than the global distribution of ID and csID features, which contribute to resilience against interference from covariate shift.}
\vspace{-3pt}
\label{fig:UMAP2}
\end{figure}

To provide stronger evidence for the motivation of our proposed method, we visualize the `goose' feature from ID (ImageNet-1K) and csID (ImageNet-C, ImageNet-R and ImageNet-V2) data. As shown in \cref{fig:UMAP2}, the global distribution of csID features differs from that of ID features owing to the covariate shift from the diverse and complex domains of csID. However, the high-likelihood ID and csID features share a compact and consistent distribution regardless of the interference of covariate shift. These high-likelihood features concentrated at class centers (\textit{i.e.}, the feature means) maintain the semantic consistency and assist in overcoming covariate shift.

\section{More Analysis of ID Semantic-Relevant Regions}
We explore the effectiveness of the sampled ID semantic-relevant regions $\boldsymbol{h}$ in other prompt tuning methods including CoOp \cite{coop} and LoCoOp \cite{miyai2023locoop} here. As shown in \cref{T:h}, $\boldsymbol{h}$ improves all results of CoOp and LoCoOp on the OOD detection metrics, again indicating the efficacy of $\boldsymbol{h}$ to promote models to overcome the interference of covariate shift and enhance OOD detection performance. It is worth noting that the enhancement of $\boldsymbol{h}$ to Far-OOD is significantly weaker than that to Near-OOD, in both CoOp and LoCoOp. This phenomenon is consistent with the analysis in Tab.~\textcolor{red}{3} of our main text, which suggests that $\boldsymbol{h}$ can purely reflect the attributes of semantics and will significantly enhance the detection performance on Near-OOD which can't be easily detected considering its slight covariate shift, while the distinct covariance information from global images can improve the detection to Far-OOD.

Moreover, the advantages of ID semantic-relevant regions $\boldsymbol{h}$ sampled from the class-conditional feature Gaussians are: \textbf{1)} Sampling semantically high-likelihood regions from the low-dimensional feature space is more tractable than from pixel or patch level, especially when a fixed number of patches need to be chosen but the sizes of foreground objects (semantic-relevant regions) vary. \textbf{2)} We can flexibly select desired regions from the Gaussians based on their probability density. \textbf{3)} These sampled regions can directly participate in training with the same dimensions as image embeddings.

\begin{table}[btp!]
\centering
\setlength{\tabcolsep}{1.5mm}
{
\caption{\textbf{Effect of ID semantic-relevant regions $\boldsymbol{h}$ in other methods.} `+$\boldsymbol{h}$' denotes that we introduce $\boldsymbol{h}$ into the cross-entropy losses employed in CoOp \cite{coop} and LoCoOp \cite{miyai2023locoop}. `Near'/`Far' means `Near-OOD'/`Far-OOD'. $\uparrow$/$\downarrow$ indicates higher/lower value is better. \textbf{$\vartriangle$} is the subtraction results between with/without $\boldsymbol{h}$ during training. The best results are in \textbf{bold}}
\label{T:h}
\small
\begin{tabular}{c|cc|cc|cc}
\toprule 
 & \multicolumn{2}{c|}{FPR@95~$\downarrow$}   & \multicolumn{2}{c|}{AUROC~$\uparrow$}   & \multicolumn{2}{c}{AUPR-IN~$\uparrow$}    \\      \cmidrule(lr){2-7}
\multirow{-2}{*}{Strategy}  & Near  & Far    & Near   & Far  & Near   & Far   \\   \cmidrule(lr){1-7}
CoOp &   \cellcolor[HTML]{E6ECE3}91.84      & \cellcolor[HTML]{ffefe0}67.60      & \cellcolor[HTML]{E6ECE3}63.69      & \cellcolor[HTML]{ffefe0}85.75 & \cellcolor[HTML]{E6ECE3}76.11 & \cellcolor[HTML]{ffefe0}95.92 \\
CoOp+$\boldsymbol{h}$ & \cellcolor[HTML]{E6ECE3}\textbf{75.94}      & \cellcolor[HTML]{ffefe0}\textbf{65.22}      & \cellcolor[HTML]{E6ECE3}\textbf{72.07}      & \cellcolor[HTML]{ffefe0}\textbf{85.82} & \cellcolor[HTML]{E6ECE3}\textbf{80.26} & \cellcolor[HTML]{ffefe0}\textbf{96.34} \\
\textbf{$\vartriangle$} & \cellcolor[HTML]{E6ECE3}-15.90      & \cellcolor[HTML]{ffefe0}-2.38     & \cellcolor[HTML]{E6ECE3}+8.38      & \cellcolor[HTML]{ffefe0}+0.07 & \cellcolor[HTML]{E6ECE3}+4.15 & \cellcolor[HTML]{ffefe0}+0.42 \\
\midrule
LoCoOp &   \cellcolor[HTML]{E6ECE3}81.18      & \cellcolor[HTML]{ffefe0}49.82      & \cellcolor[HTML]{E6ECE3}70.15      & \cellcolor[HTML]{ffefe0}89.10 & \cellcolor[HTML]{E6ECE3}79.16 & \cellcolor[HTML]{ffefe0}96.76 \\
LoCoOp+$\boldsymbol{h}$ & \cellcolor[HTML]{E6ECE3}\textbf{75.73}      & \cellcolor[HTML]{ffefe0}\textbf{48.47}      & \cellcolor[HTML]{E6ECE3}\textbf{74.84}      & \cellcolor[HTML]{ffefe0}\textbf{89.51} & \cellcolor[HTML]{E6ECE3}\textbf{82.35} & \cellcolor[HTML]{ffefe0}\textbf{97.24} \\
\textbf{$\vartriangle$} & \cellcolor[HTML]{E6ECE3}-5.45      & \cellcolor[HTML]{ffefe0}-1.35     & \cellcolor[HTML]{E6ECE3}+4.69      & \cellcolor[HTML]{ffefe0}+0.41 & \cellcolor[HTML]{E6ECE3}+3.19 & \cellcolor[HTML]{ffefe0}+0.48 \\
\bottomrule
\end{tabular}}
\vspace{10pt}
\end{table}

\begin{table}[btp!]
\centering
\small
\setlength{\tabcolsep}{1.0mm}
{
\caption{\textbf{Effect of the expanded learnable OOD context in LSA.} Our proposed LSA tunes the CLIP with OOD context, and the training object is written as \cref{e:noOODloss} when deprecating OOD context. `Near'/`Far' means `Near-OOD'/`Far-OOD'. The best results are in \textbf{bold}.}
\label{T:OODcontext}
\begin{tabular}{c|cc|cc|cc}
\toprule 
 & \multicolumn{2}{c|}{FPR@95~$\downarrow$}   & \multicolumn{2}{c|}{AUROC~$\uparrow$}   & \multicolumn{2}{c}{AUPR-IN~$\uparrow$}    \\      \cmidrule(lr){2-7}
\multirow{-2}{*}{Strategy}  & Near  & Far    & Near   & Far  & Near   & Far   \\   \cmidrule(lr){1-7}
\textit{w/o} OOD context &   \cellcolor[HTML]{E6ECE3}68.63 & \cellcolor[HTML]{ffefe0}48.24 & \cellcolor[HTML]{E6ECE3}79.30 & \cellcolor[HTML]{ffefe0}89.97 & \cellcolor[HTML]{E6ECE3}84.10 & \cellcolor[HTML]{ffefe0}97.36 \\
\textit{w} OOD context & \cellcolor[HTML]{E6ECE3}\textbf{52.73}      & \cellcolor[HTML]{ffefe0}\textbf{32.41}      & \cellcolor[HTML]{E6ECE3}\textbf{85.41}      & \cellcolor[HTML]{ffefe0}\textbf{90.97} & \cellcolor[HTML]{E6ECE3}\textbf{90.26} & \cellcolor[HTML]{ffefe0}\textbf{97.49} \\
\bottomrule
\end{tabular}}
\vspace{-2pt}
\end{table}

\section{Effect of Learnable OOD Context}
For the text branch of our proposed LSA, we design the bidirectional prompt customization mechanism which expands the additional learnable OOD (negative) context to enlarge the discrepancy between ID and OOD semantics in text space. Here we analyze the effect of the expanded OOD context. Specifically, we abandon the OOD context in LSA and accordingly reconstructed the training objective as:
\begin{equation}\label{e:noOODloss}
\mathcal{L}^{\prime} = \mathcal{L}_{ce}^{\prime} + \gamma \mathcal{L}_{uni}^{\prime},
\end{equation}
where $\gamma$ is still set as $0.5$, $\mathcal{L}_{ce}^{\prime}$ and $\mathcal{L}_{uni}^{\prime}$ are formulated as \cref{e:noOODceloss} and \cref{e:noOODuniloss}, respectively.
\begin{equation}\label{e:noOODceloss}
\mathcal{L}_{ce}^{\prime} = -\mathbb{E}_{(\boldsymbol{v}^{\prime},y)\sim \mathcal{V}^{\prime}}[\ell_{\mathrm{CE}}(p(\boldsymbol{y}|\boldsymbol{v}^{\prime}),y)],
\end{equation}
\begin{equation}\label{e:noOODuniloss}
\mathcal{L}_{uni}^{\prime} =-\mathbb{E}_{\boldsymbol{o}\sim \mathcal{O}}[\ell_{\mathrm{OE}}(p(\boldsymbol{y_{id}}|\boldsymbol{o}))],
\end{equation}
where $\mathcal{V}^{\prime} = \{\boldsymbol{\varphi}_v(x_{i^{\prime}})|x_{i^{\prime}}\in\mathcal{D}_{\text{few}}\} \cup \mathcal{H}_s$, and $p(\boldsymbol{y_{id}}|\boldsymbol{o})$ is the similarity of OOD regularization regions $\boldsymbol{o}$ with all ID contexts. The definitions of $\mathcal{D}_{\text{few}}$, $\mathcal{H}_s$ and $\ell_{\mathrm{OE}}$ are the same as that in main text. 

As the comparison in \cref{T:OODcontext} shows, expanding OOD context during optimization significantly enhances the performance of LSA. It indicates that in addition to exploiting the feature of pure semantic regions, it is also essential to design appropriate methods to explicitly learn the discrepancy between ID/OOD semantics for OOD detection. Note that LSA can still achieve state-of-the-art performance even without using learnable OOD context.

\begin{table}[btp!]
\centering
\setlength{\tabcolsep}{1.5mm}
{
\caption{\textbf{Effect of label assignment to OOD regularization regions $\boldsymbol{o}$.} `OT' denotes the label assignment method with Optimal Transport used in SeLa \cite{sela} and `Random' means the random assignment of labels to $\boldsymbol{o}$. LSA applies the latter one. `Near'/`Far' means `Near-OOD'/`Far-OOD'. The best results are in \textbf{bold}.}
\label{T:labelassignment}
\begin{tabular}{c|cc|cc|cc}
\toprule 
 & \multicolumn{2}{c|}{FPR@95~$\downarrow$}   & \multicolumn{2}{c|}{AUROC~$\uparrow$}   & \multicolumn{2}{c}{AUPR-IN~$\uparrow$}    \\      \cmidrule(lr){2-7}
\multirow{-2}{*}{Strategy}  & Near  & Far    & Near   & Far  & Near   & Far   \\   \cmidrule(lr){1-7}
OT &   \cellcolor[HTML]{E6ECE3}55.72 & \cellcolor[HTML]{ffefe0}34.84 & \cellcolor[HTML]{E6ECE3}82.80 & \cellcolor[HTML]{ffefe0}89.80 & \cellcolor[HTML]{E6ECE3}89.30 & \cellcolor[HTML]{ffefe0}96.86 \\
Random & \cellcolor[HTML]{E6ECE3}\textbf{52.73}      & \cellcolor[HTML]{ffefe0}\textbf{32.41}      & \cellcolor[HTML]{E6ECE3}\textbf{85.41}      & \cellcolor[HTML]{ffefe0}\textbf{90.97} & \cellcolor[HTML]{E6ECE3}\textbf{90.26} & \cellcolor[HTML]{ffefe0}\textbf{97.49} \\
\bottomrule
\end{tabular}}
\end{table}

\begin{table}[btp!]
\centering
\renewcommand{\arraystretch}{1.}
\renewcommand{\tabcolsep}{1.pt}
\small
\caption{\textbf{Effect of different designs of learnable ID context.} `Unified' means sharing the same learnable context with all ID classes, while `CSC' denotes learning class-specific context. Our proposed LSA employ `CSC' during optimization. The best results are in \textbf{bold} and the second best are \underline{underlined}.}
\label{T:CSC}
\begin{tabular}{c|cc|cc|c}
\toprule 
& \multicolumn{2}{c|}{FPR@95~$\downarrow$}         & \multicolumn{2}{c|}{AUROC~$\uparrow$}    \\      \cmidrule(lr){2-5}
\multirow{-2}{*}{\normalsize{Methods}}  & \small{Near-OOD}   & \small{Far-OOD}    & \small{Near-OOD}   & \small{Far-OOD}  & \multirow{-2}{*}{ACC~$\uparrow$}  \\   \cmidrule(lr){1-6}
MCM \cite{mcm} &          \cellcolor[HTML]{E6ECE3}87.24 & \cellcolor[HTML]{ffefe0}62.43& \cellcolor[HTML]{E6ECE3}70.07 & \cellcolor[HTML]{ffefe0}88.99 & 76.37 \\
CoOp \cite{coop}                         & \cellcolor[HTML]{E6ECE3}91.84      & \cellcolor[HTML]{ffefe0}67.60      & \cellcolor[HTML]{E6ECE3}63.69      & \cellcolor[HTML]{ffefe0}85.75 &        73.41              \\
CoCoOp \cite{cocoop}                         & \cellcolor[HTML]{E6ECE3}89.07      & \cellcolor[HTML]{ffefe0}70.29      & \cellcolor[HTML]{E6ECE3}67.17      & \cellcolor[HTML]{ffefe0}88.42 &         74.32              \\
MaPLe \cite{khattak2023maple}                         & \cellcolor[HTML]{E6ECE3}91.19      & \cellcolor[HTML]{ffefe0}72.72      & \cellcolor[HTML]{E6ECE3}67.28      & \cellcolor[HTML]{ffefe0}86.19 &          73.17           \\
LoCoOp \cite{miyai2023locoop} &   \cellcolor[HTML]{E6ECE3}81.18 & \cellcolor[HTML]{ffefe0}49.82      & \cellcolor[HTML]{E6ECE3}70.15     & \cellcolor[HTML]{ffefe0}89.10 & 75.33 \\
\textbf{Unified (Ours)} &  \cellcolor[HTML]{E6ECE3}\underline{61.67} & \cellcolor[HTML]{ffefe0}\underline{45.81}     & \cellcolor[HTML]{E6ECE3}\underline{79.25}      & \cellcolor[HTML]{ffefe0}\underline{89.33} & \textbf{77.11} \\
\textbf{CSC (Ours)} &  \cellcolor[HTML]{E6ECE3}\textbf{52.73} & \cellcolor[HTML]{ffefe0}\textbf{32.41}      & \cellcolor[HTML]{E6ECE3}\textbf{85.41}      & \cellcolor[HTML]{ffefe0}\textbf{90.97} & \underline{76.54} \\

\bottomrule
\end{tabular}
\vspace{-4pt}
\end{table}

\section{Label Assignment to OOD Regularization Regions}
Considering we simply assign a random OOD label to each OOD regularization region $\boldsymbol{o}$ in the CE loss $\mathcal{L}_{ce}$, we apply the label assignment based on Optimal Transport (OT) like SeLa \cite{sela} and analyze the effect of different label assignment methods here. Following SeLa \cite{sela}, we define the similarities of a batch of $\boldsymbol{o}$ with all $M$ OOD contexts as the cost matrix $\mathbf{P}\in\mathbb{R}^{M\times B}$ in each iteration, where $B$ is the number of a batch of $\boldsymbol{o}$. Then we implement uniform prior constraint on the label assignment matrix and introduce a Kullback-Leibler divergence regularization term to adopt a faster Sinkhorn-Knopp algorithm \cite{cuturi2013sinkhorn}. Finally, we obtain the OOD label assignment matrix $\mathbf{Q}\in\mathbb{R}^{M\times B}$ for all $\boldsymbol{o}$ in each iteration. As shown in \cref{T:labelassignment}, the performance with label assignment based on OT is also excellent and its gap with random label assignment is slight. The purpose of the OT method applied in SeLa \cite{sela} is to unify the probability distribution of all classes and avoid degenerate solutions, which can be easily achieved by assigning labels randomly, especially when OOD regularization regions do not contain explicit semantic information.

\section{Design of the Learnable ID Context}
The class-specific context (CSC) learned in our method needs to optimize more text embeddings than unified context (Unified) for all classes, and we explore the effect of applying Unified here. As shown in \cref{T:CSC}, our method outperforms all CLIP-based prompt tuning and OOD detection methods when employing the learnable unified context, while it is inferior to using CSC. This is because CSC takes into account the diversity of ID semantic classes and the difference between ID objects. Moreover, our method with Unified achieves the best in classification, as pointed out in CoOp \cite{coop} that using unified context is better than CSC for classifying images with generic objects, such as ImageNet and its derived datasets.

\begin{table}[btp!]
\centering
\setlength{\tabcolsep}{1.5mm}
{
\small
\caption{\textbf{Comparison between LSA and CLIPN \cite{wang2023clipn}.} $\uparrow$/$\downarrow$ indicates higher/lower value is better. `Near'/`Far' means `Near-OOD'/`Far-OOD'. The best results are in \textbf{bold}.}
\label{T:withCLIPN}
\begin{tabular}{c|cc|cc|cc}
\toprule 
 \multicolumn{7}{c}{\textbf{Benchmark: ImageNet-200}}         \\  \cmidrule(lr){1-7}
& \multicolumn{2}{c|}{\footnotesize{FPR@95~$\downarrow$}}    & \multicolumn{2}{c|}{\footnotesize{AUROC~$\uparrow$}}   & \multicolumn{2}{c}{\footnotesize{AUPR-IN~$\uparrow$}}    \\      \cmidrule(lr){2-7}
\multirow{-2}{*}{Method}  & \small{Near}   & \small{Far}    & \small{Near}   & \small{Far}  & \small{Near}   & \small{Far}   \\   \cmidrule(lr){1-7}
CLIPN \cite{wang2023clipn} &   \cellcolor[HTML]{E6ECE3}57.84 & \cellcolor[HTML]{ffefe0}\textbf{25.82} & \cellcolor[HTML]{E6ECE3}79.45 & \cellcolor[HTML]{ffefe0}\textbf{93.49} & \cellcolor[HTML]{E6ECE3}89.16 & \cellcolor[HTML]{ffefe0}\textbf{98.28} \\
\textbf{LSA(ours)} & \cellcolor[HTML]{E6ECE3}\textbf{52.73}      & \cellcolor[HTML]{ffefe0}32.41      & \cellcolor[HTML]{E6ECE3}\textbf{85.41}      & \cellcolor[HTML]{ffefe0}90.97 & \cellcolor[HTML]{E6ECE3}\textbf{90.26} & \cellcolor[HTML]{ffefe0}97.49 \\
\bottomrule
\end{tabular}

\begin{tabular}{c|cc|cc|cc}
\toprule 
 \multicolumn{7}{c}{\textbf{Benchmark: ImageNet-1K}}         \\  \cmidrule(lr){1-7}
& \multicolumn{2}{c|}{\footnotesize{FPR@95~$\downarrow$}}    & \multicolumn{2}{c|}{\footnotesize{AUROC~$\uparrow$}}   & \multicolumn{2}{c}{\footnotesize{AUPR-IN~$\uparrow$}}    \\      \cmidrule(lr){2-7}
\multirow{-2}{*}{Method}  & \small{Near}   & \small{Far}    & \small{Near}   & \small{Far}  & \small{Near}   & \small{Far}   \\   \cmidrule(lr){1-7}
CLIPN \cite{wang2023clipn} &  \cellcolor[HTML]{E6ECE3}75.40 & \cellcolor[HTML]{ffefe0}\textbf{46.14} & \cellcolor[HTML]{E6ECE3}71.56 & \cellcolor[HTML]{ffefe0}\textbf{90.24} & \cellcolor[HTML]{E6ECE3}90.96 & \cellcolor[HTML]{ffefe0}\textbf{98.34} \\
\textbf{LSA(ours)} & \cellcolor[HTML]{E6ECE3}\textbf{70.56}      & \cellcolor[HTML]{ffefe0}48.06      & \cellcolor[HTML]{E6ECE3}\textbf{78.22}      & \cellcolor[HTML]{ffefe0}86.85 & \cellcolor[HTML]{E6ECE3}\textbf{93.05} & \cellcolor[HTML]{ffefe0}98.10 \\
\bottomrule
\end{tabular}}
\end{table}

\begin{table}[btp!]
\centering
\setlength{\tabcolsep}{1.5mm}
{
\small
\caption{\textbf{Comparison between LSA+ and CLIPN \cite{wang2023clipn}.} LSA+ introduces the extra few-shot outlier data into training based on our proposed LSA. `Near'/`Far' means `Near-OOD'/`Far-OOD'. The best results are in \textbf{bold}.}
\label{T:OEwithCLIPN}
\begin{tabular}{c|cc|cc|cc}
\toprule 
 \multicolumn{7}{c}{\textbf{Benchmark: ImageNet-200}}         \\  \cmidrule(lr){1-7}
& \multicolumn{2}{c|}{\footnotesize{FPR@95~$\downarrow$}}    & \multicolumn{2}{c|}{\footnotesize{AUROC~$\uparrow$}}   & \multicolumn{2}{c}{\footnotesize{AUPR-IN~$\uparrow$}}    \\      \cmidrule(lr){2-7}
\multirow{-2}{*}{Method}  & \small{Near}   & \small{Far}    & \small{Near}   & \small{Far}  & \small{Near}   & \small{Far}   \\   \cmidrule(lr){1-7}
CLIPN \cite{wang2023clipn} &   \cellcolor[HTML]{E6ECE3}57.84 & \cellcolor[HTML]{ffefe0}25.82 & \cellcolor[HTML]{E6ECE3}79.45 & \cellcolor[HTML]{ffefe0}93.49 & \cellcolor[HTML]{E6ECE3}89.16 & \cellcolor[HTML]{ffefe0}98.28 \\
\textbf{LSA+(ours)} & \cellcolor[HTML]{E6ECE3}\textbf{47.71} & \cellcolor[HTML]{ffefe0}\textbf{22.64}      & \cellcolor[HTML]{E6ECE3}\textbf{88.10}     & \cellcolor[HTML]{ffefe0}\textbf{95.16}  & \cellcolor[HTML]{E6ECE3}\textbf{91.28} & \cellcolor[HTML]{ffefe0}\textbf{98.86} \\
\bottomrule
\end{tabular}

\begin{tabular}{c|cc|cc|cc}
\toprule 
 \multicolumn{7}{c}{\textbf{Benchmark: ImageNet-1K}}         \\  \cmidrule(lr){1-7}
& \multicolumn{2}{c|}{\footnotesize{FPR@95~$\downarrow$}}    & \multicolumn{2}{c|}{\footnotesize{AUROC~$\uparrow$}}   & \multicolumn{2}{c}{\footnotesize{AUPR-IN~$\uparrow$}}    \\      \cmidrule(lr){2-7}
\multirow{-2}{*}{Method}  & \small{Near}   & \small{Far}    & \small{Near}   & \small{Far}  & \small{Near}   & \small{Far}   \\   \cmidrule(lr){1-7}
CLIPN \cite{wang2023clipn} &  \cellcolor[HTML]{E6ECE3}75.40 & \cellcolor[HTML]{ffefe0}46.14 & \cellcolor[HTML]{E6ECE3}71.56 & \cellcolor[HTML]{ffefe0}90.24 & \cellcolor[HTML]{E6ECE3}90.96 & \cellcolor[HTML]{ffefe0}98.34 \\
\textbf{LSA+(ours)} & \cellcolor[HTML]{E6ECE3}\textbf{67.47} & \cellcolor[HTML]{ffefe0}\textbf{43.82}      & \cellcolor[HTML]{E6ECE3}\textbf{79.94}     & \cellcolor[HTML]{ffefe0}\textbf{91.23}  & \cellcolor[HTML]{E6ECE3}\textbf{93.92} & \cellcolor[HTML]{ffefe0}\textbf{98.85} \\
\bottomrule
\end{tabular}}
\end{table}

\begin{table*}[btp!]
\renewcommand{\arraystretch}{1.}
\renewcommand{\tabcolsep}{8.pt}
\centering
\caption{\textbf{The dataset-specific results of our proposed LSA on two F-OOD benchmarks.} $\uparrow$/$\downarrow$ indicates higher/lower value is better, and the average results of Near-OOD and Far-OOD are denoted as `Mean' and in \textbf{bold}.}
\label{table:datasetResults}
\begin{tabular}{cc|c|c|c|c}
\toprule 
                          \multicolumn{6}{c}{\textbf{Benchmark: ImageNet-200}}         \\  \cmidrule(lr){1-6}
Near-/Far-OOD  &  Dataset                        & \multicolumn{1}{c|}{FPR@95~$\downarrow$}     & \multicolumn{1}{c|}{AUROC~$\uparrow$}                               & \multicolumn{1}{c|}{AUPR-IN~$\uparrow$}             & \multicolumn{1}{c}{AUPR-OUT~$\uparrow$}                      \\ \cmidrule(lr){1-6}
\multirow{3}{*}{\begin{tabular}[c]{@{}c@{}}Near-OOD\end{tabular}} 
&  SSB-hard \cite{SSB}       & \cellcolor[HTML]{E6ECE3}49.84 & \cellcolor[HTML]{E6ECE3}85.63 & \cellcolor[HTML]{E6ECE3}83.20 & \cellcolor[HTML]{E6ECE3}87.86         \\
& NINCO \cite{ninco} & \cellcolor[HTML]{E6ECE3}55.62 & \cellcolor[HTML]{E6ECE3}85.19 & \cellcolor[HTML]{E6ECE3}97.32 & \cellcolor[HTML]{E6ECE3}51.29         \\
& \textbf{Mean}  & \cellcolor[HTML]{E6ECE3}\textbf{52.73} & \cellcolor[HTML]{E6ECE3}\textbf{85.41} & \cellcolor[HTML]{E6ECE3}\textbf{90.26} & \cellcolor[HTML]{E6ECE3}\textbf{69.58}         \\  
\midrule
\multirow{4}{*}{\begin{tabular}[c]{@{}c@{}}Far-OOD\end{tabular}} 
&  iNaturalist \cite{inaturalist}       & \cellcolor[HTML]{ffefe0}12.34 & \cellcolor[HTML]{ffefe0}97.30 & \cellcolor[HTML]{ffefe0}99.29 & \cellcolor[HTML]{ffefe0}91.32         \\
& Textures \cite{texture} & \cellcolor[HTML]{ffefe0}52.58 & \cellcolor[HTML]{ffefe0}84.18 & \cellcolor[HTML]{ffefe0}97.29 & \cellcolor[HTML]{ffefe0}48.78         \\
& OpenImage-O \cite{haoqi2022vim} & \cellcolor[HTML]{ffefe0}32.31 & \cellcolor[HTML]{ffefe0}91.43 & \cellcolor[HTML]{ffefe0}95.89 & \cellcolor[HTML]{ffefe0}84.66         \\
& \textbf{Mean}  & \cellcolor[HTML]{ffefe0}\textbf{32.41} & \cellcolor[HTML]{ffefe0}\textbf{90.97} & \cellcolor[HTML]{ffefe0}\textbf{97.49} & \cellcolor[HTML]{ffefe0}\textbf{74.92}         \\ 
\bottomrule
\end{tabular} 

\begin{tabular}{cc|c|c|c|c}
\toprule 
                          \multicolumn{6}{c}{\textbf{Benchmark: ImageNet-1K}}         \\  \cmidrule(lr){1-6}
Near-/Far-OOD  &  Dataset                        & \multicolumn{1}{c|}{FPR@95~$\downarrow$}     & \multicolumn{1}{c|}{AUROC~$\uparrow$}                               & \multicolumn{1}{c|}{AUPR-IN~$\uparrow$}             & \multicolumn{1}{c}{AUPR-OUT~$\uparrow$}                      \\ \cmidrule(lr){1-6}
\multirow{3}{*}{\begin{tabular}[c]{@{}c@{}}Near-OOD\end{tabular}} 
&  SSB-hard \cite{SSB}       & \cellcolor[HTML]{E6ECE3}69.13 & \cellcolor[HTML]{E6ECE3}79.57 & \cellcolor[HTML]{E6ECE3}88.11 & \cellcolor[HTML]{E6ECE3}66.72         \\
& NINCO \cite{ninco} & \cellcolor[HTML]{E6ECE3}72.00 & \cellcolor[HTML]{E6ECE3}76.88 & \cellcolor[HTML]{E6ECE3}97.98 & \cellcolor[HTML]{E6ECE3}20.34         \\
& \textbf{Mean}  & \cellcolor[HTML]{E6ECE3}\textbf{70.56} & \cellcolor[HTML]{E6ECE3}\textbf{78.22}  & \cellcolor[HTML]{E6ECE3}\textbf{93.05} & \cellcolor[HTML]{E6ECE3}\textbf{43.53}         \\  
\midrule
\multirow{4}{*}{\begin{tabular}[c]{@{}c@{}}Far-OOD\end{tabular}} 
&  iNaturalist \cite{inaturalist}       & \cellcolor[HTML]{ffefe0}28.53 & \cellcolor[HTML]{ffefe0}93.90 & \cellcolor[HTML]{ffefe0}99.24 & \cellcolor[HTML]{ffefe0}72.35         \\
& Textures \cite{texture} & \cellcolor[HTML]{ffefe0}65.21 & \cellcolor[HTML]{ffefe0}79.41 & \cellcolor[HTML]{ffefe0}98.39 & \cellcolor[HTML]{ffefe0}24.18         \\
& OpenImage-O \cite{haoqi2022vim} & \cellcolor[HTML]{ffefe0}50.44 & \cellcolor[HTML]{ffefe0}85.44 & \cellcolor[HTML]{ffefe0}96.62 & \cellcolor[HTML]{ffefe0}61.01         \\
& \textbf{Mean}  & \cellcolor[HTML]{ffefe0}\textbf{48.06} & \cellcolor[HTML]{ffefe0}\textbf{86.85} & \cellcolor[HTML]{ffefe0}\textbf{98.10} & \cellcolor[HTML]{ffefe0}\textbf{52.51}         \\ 
\bottomrule
\end{tabular} 
\vspace{10pt}
\end{table*}

\begin{table*}[btp!]
\renewcommand{\arraystretch}{1.}
\renewcommand{\tabcolsep}{8.pt}
\centering
\caption{\textbf{Effect of visual encoder backbones.} $\uparrow$/$\downarrow$ indicates higher/lower value is better. The best results of CNN (ResNet-50 \& ResNet-101) and Vision Transformer (ViT-B/16 \& ViT-B/32) are in \textbf{bold}. Our proposed LSA employs the \textcolor{red}{ViT-B/16} following \cite{zeroMCMlocal,miyai2023locoop,coop} and the performance with \textcolor{red}{ViT-B/16} achieves the best on two F-OOD benchmarks.}
\label{table:backbone}
\begin{tabular}{c|cc|cc|cc}
\toprule 
\multicolumn{7}{c}{\textbf{Benchmark: ImageNet-200}}         \\  \cmidrule(lr){1-7}
& \multicolumn{2}{c|}{FPR@95~$\downarrow$}                                   & \multicolumn{2}{c|}{AUROC~$\uparrow$}                               & \multicolumn{2}{c}{AUPR-IN~$\uparrow$}                                   \\ \cmidrule(lr){2-7}
\multirow{-2}{*}{Backbones} & Near-OOD                      & Far-OOD                       & Near-OOD                      & Far-OOD                  & Near-OOD                 & Far-OOD                   \\ \cmidrule(lr){1-7}
ResNet-50                     & \cellcolor[HTML]{E6ECE3}\textbf{70.32} & \cellcolor[HTML]{ffefe0}\textbf{54.32} & \cellcolor[HTML]{E6ECE3}\textbf{74.97} & \cellcolor[HTML]{ffefe0}\textbf{82.84} & \cellcolor[HTML]{E6ECE3}\textbf{85.31} & \cellcolor[HTML]{ffefe0}\textbf{94.94}             \\
ResNet-101                 & \cellcolor[HTML]{E6ECE3}71.33 & \cellcolor[HTML]{ffefe0}60.06      & \cellcolor[HTML]{E6ECE3}72.77      & \cellcolor[HTML]{ffefe0}81.18 & \cellcolor[HTML]{E6ECE3}83.80 & \cellcolor[HTML]{ffefe0}94.33       \\
\midrule
\textcolor{red}{ViT-B/16}                         & \cellcolor[HTML]{E6ECE3}\textbf{52.73}      & \cellcolor[HTML]{ffefe0}\textbf{32.41}      & \cellcolor[HTML]{E6ECE3}\textbf{85.41}      & \cellcolor[HTML]{ffefe0}\textbf{90.97} & \cellcolor[HTML]{E6ECE3}\textbf{90.26} & \cellcolor[HTML]{ffefe0}\textbf{97.49}        \\
ViT-B/32                          & \cellcolor[HTML]{E6ECE3}59.01      & \cellcolor[HTML]{ffefe0}41.59      & \cellcolor[HTML]{E6ECE3}75.73      & \cellcolor[HTML]{ffefe0}85.57 & \cellcolor[HTML]{E6ECE3}89.22  & \cellcolor[HTML]{ffefe0}96.16 \\     

\bottomrule
\end{tabular} 
\begin{tabular}{c|cc|cc|cc}
\toprule
\multicolumn{7}{c}{\textbf{Benchmark: ImageNet-1K}}         \\  \cmidrule(lr){1-7}
& \multicolumn{2}{c|}{FPR@95~$\downarrow$}                                   & \multicolumn{2}{c|}{AUROC~$\uparrow$}                               & \multicolumn{2}{c}{AUPR-IN~$\uparrow$}                             \\ \cmidrule(lr){2-7}
\multirow{-2}{*}{Backbones} & Near-OOD                      & Far-OOD                       & Near-OOD                      & Far-OOD                  & Near-OOD                 & Far-OOD                 \\ \cmidrule(lr){1-7}
ResNet-50                     & \cellcolor[HTML]{E6ECE3}\textbf{79.80} & \cellcolor[HTML]{ffefe0}\textbf{57.16} & \cellcolor[HTML]{E6ECE3}\textbf{72.85} & \cellcolor[HTML]{ffefe0}\textbf{83.27} & \cellcolor[HTML]{E6ECE3}\textbf{90.45} & \cellcolor[HTML]{ffefe0}\textbf{97.74}        \\
ResNet-101     & \cellcolor[HTML]{E6ECE3}80.45 & \cellcolor[HTML]{ffefe0}60.22      & \cellcolor[HTML]{E6ECE3}71.46      & \cellcolor[HTML]{ffefe0}81.77 & \cellcolor[HTML]{E6ECE3}87.93 & \cellcolor[HTML]{ffefe0}96.41      \\
\midrule
\textcolor{red}{ViT-B/16}     & \cellcolor[HTML]{E6ECE3}\textbf{70.56}      & \cellcolor[HTML]{ffefe0}\textbf{48.06}      & \cellcolor[HTML]{E6ECE3}\textbf{78.22}      & \cellcolor[HTML]{ffefe0}\textbf{86.85} & \cellcolor[HTML]{E6ECE3}\textbf{93.05} & \cellcolor[HTML]{ffefe0}\textbf{98.10}  \\
ViT-B/32                         & \cellcolor[HTML]{E6ECE3}76.87      & \cellcolor[HTML]{ffefe0}54.81      & \cellcolor[HTML]{E6ECE3}73.28      & \cellcolor[HTML]{ffefe0}83.01   & \cellcolor[HTML]{E6ECE3}90.78 & \cellcolor[HTML]{ffefe0}97.59        \\

\bottomrule
\end{tabular}
\vspace{3pt}
\end{table*}

\section{Performance Comparison with CLIPN}
We notice a recent CLIP-based OOD detection work CLIPN \cite{wang2023clipn}, which empowers the logic of saying `no' within CLIP utilizing the large-scale image-text paired dataset, CC-3M \cite{cc}. The reason why we did not compare performance with CLIPN in the main text is that, CLIPN does not comply with the training protocol of F-OOD task. F-OOD protocol in OpenOODv1.5 \cite{zhang2023openood} suggests that ImageNet-200 benchmark can access ImageNet-200 training set during training and we also take the rest 800 classes' images from ImageNet-1K as the outlier training set for it, while only ImageNet-1K training set is available to ImageNet-1K benchmark considering it is difficult to find outlier training images do not overlap with the corresponding OOD testing set. Hence, according to the F-OOD protocol, CLIPN exploits massive non-compliant outlier data from CC-3M during training. We compare the performance between our proposed LSA and CLIPN in \cref{T:withCLIPN} and demonstrate the superiority of LSA in detecting the intractable Near-OOD. 

Then we randomly select 800 classes from ImageNet-21K that do not overlap with the 1000 categories in ImageNet-1K and utilize 16 images from each of these 800 classes to construct the few-shot outlier training set $\mathcal{D}_{\text{OOD}}^{few}$ for both ImageNet-200 benchmark and ImageNet-1K benchmark. We replace the OOD regularization regions in our training object with the embeddings from $\mathcal{D}_{\text{OOD}}^{few}$ and the label set of $\mathcal{D}_{\text{OOD}}^{few}$ is $\mathcal{Y}_{OOD} = \{C,C+1,...,C+799\}$, where $C$ is the number of ID classes. Correspondingly, the learnable OOD textual vectors are denoted as $\boldsymbol{T^{ood}} \in \mathbb{R}^{800 \times D}$, where $D$ is the dimension of text embedding output by the text encoder of CLIP. We present the performance of LSA employing the extra $\mathcal{D}_{\text{OOD}}^{few}$ and compare it with CLIPN in \cref{T:OEwithCLIPN}. The comparison shows that LSA comprehensively outperforms CLIPN when only a tiny amount of outlier images is introduced into training.

\begin{figure}[t]
\small
\centering
\begin{overpic}[width=1\linewidth]{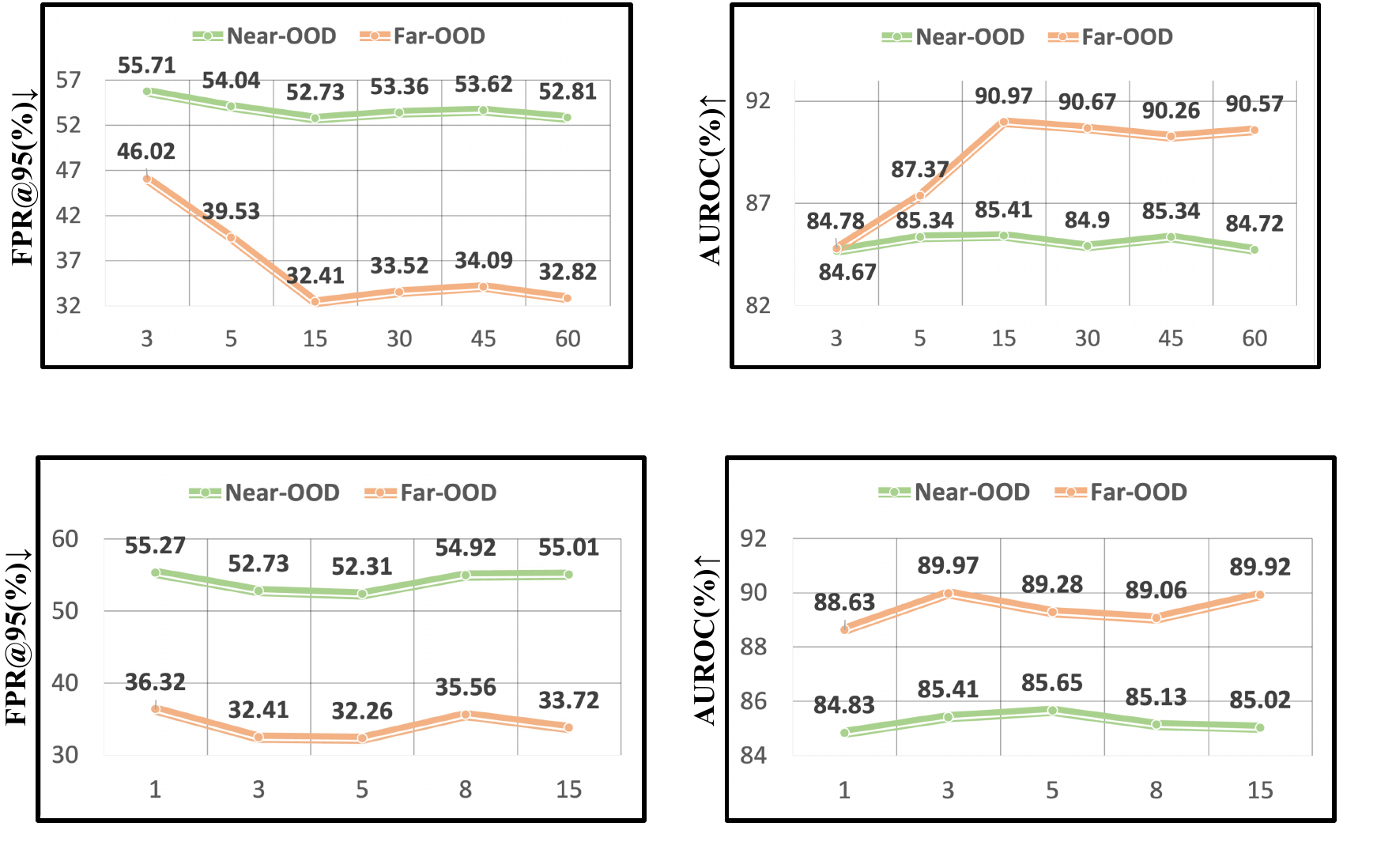}
        \put(25, 33){(a) Comparison between different $M$}
        \put(25, 0.0){(b) Comparison between different $K$}
\end{overpic}
\vspace{-8pt}
\caption{\textbf{Analysis on hyper-parameters.} We analyze the effect of the number of expanded OOD contexts $M$ and the number of tokens $K$ for each learnable ID and OOD context. $\uparrow$/$\downarrow$ indicates higher/lower value is better. (a) shows our proposed LSA can obtain a stably good result when $M\ge15$, and (b) proves that LSA is insensitive to the change of $K$.} 
\vspace{-3pt}
\label{fig:MKhyper}
\end{figure}

\section{Detailed Results and More Backbones}
In this section, we first present the detailed results among all datasets of LSA in \cref{table:datasetResults}. Far-OOD datasets are generally easier to detect than Near-OOD, and their experimental results are better. However, the empirical difficulty of Textures \cite{texture} is more significant than other Far-OOD datasets and comparable to Near-OOD as shown in \cref{table:datasetResults}. It can be explained that the Textures \cite{texture} dataset has flat backgrounds compared to ImageNet and lacks clear semantic information (\textit{i.e.}, specific foreground objects), which will be identified as OOD relying on the resulting covariate shift, but it may be not so easy for LSA that focuses on semantic shift. 

Then we compare the performance of LSA with different visual encoder backbones. In general, LSA favors Vision Transformer as the visual encoder and achieves the best results with ViT-B/ on the two F-OOD benchmarks. However, it does not mean larger-scale models will necessarily improve performance. As shown in \cref{table:backbone}, LSA obtains the best results employing ResNet-50 and ViT-B/16 with fewer parameters in terms of CNN and Vision Transformer architectures, respectively. Notably, LSA with ResNet-50 still outperforms all ResNet-based methods in Tab.\textcolor{red}{1} (the first four rows of each benchmark table) on 11 out of 12 metrics. All of these prove that employing a large-scale and strong architecture is not the key to success in OOD detection tasks, as indicated in OpenOODv1.5 \cite{zhang2023openood}.

\section{Analysis of More Hyper-parameters}\label{sec:hyperparameters}
Here we analyze the effect of hyper-parameters including the number of expanded OOD contexts $M$ and the number of tokens $K$ for each learnable ID and OOD context. \cref{fig:MKhyper}\textcolor{red}{(a)} shows that the performance of LSA on Near-OOD is consistently stable when $M$ ranges from $3$ to $60$, and the performance fluctuation on Far-OOD is slight when $M\ge15$. \cref{fig:MKhyper}\textcolor{red}{(b)} indicates that the maximum fluctuation of all metrics on both Near-OOD and Far-OOD is only about $3\%$ when $K\ge3$. All of these prove the robustness of our proposed LSA, and we choose $M=15$ and $K=3$ to optimize the least prompts while ensuring excellent performance.

\end{document}